\documentclass{article}

\usepackage{microtype}
\usepackage{graphicx}
\usepackage{subfigure}
\usepackage{booktabs} % for professional tables

\usepackage{hyperref}

\usepackage[table]{xcolor}
\usepackage[accepted]{icml2025}

\usepackage{amsmath}
\usepackage{amssymb}
\usepackage{mathtools}
\usepackage{amsthm}

\usepackage{bm}
\usepackage{enumitem}
\usepackage{url}
\usepackage{graphicx} 
\usepackage{wrapfig}
\usepackage{booktabs} 
\usepackage{multirow}
\usepackage{siunitx}
\usepackage{marvosym}

\usepackage[capitalize,noabbrev]{cleveref}

\theoremstyle{plain}

\theoremstyle{definition}

\theoremstyle{remark}

\usepackage[textsize=tiny]{todonotes}

\icmltitlerunning{Domain2Vec: Vectorizing Datasets to Find the Optimal Data Mixture without Training}

\begin{document}

\twocolumn[
\icmltitle{Domain2Vec: Vectorizing Datasets to Find the Optimal Data Mixture without Training}




\begin{icmlauthorlist}
\icmlauthor{Mozhi Zhang}{sch}
\icmlauthor{Howe Tissue \Letter}{}
\icmlauthor{Lu Wang}{comp}
\icmlauthor{Xipeng Qiu}{sch}

\end{icmlauthorlist}

\icmlaffiliation{sch}{School of Computer Science, Fudan University, Shanghai, China}
\icmlaffiliation{comp}{Ritzz-AI}

\icmlcorrespondingauthor{Howe Tissue (project lead)}{\url{h-sun20@tsinghua.org.cn}}

\icmlkeywords{Machine Learning, ICML}

\vskip 0.3in
]



\printAffiliationsAndNotice{}

\begin{abstract}
% The mixture ratio of data from different source domains significantly affects the performance of language models (LM) pretraining.
We introduce~\textsc{Domain2Vec}, a novel approach that decomposes any dataset into a linear combination of several \emph{meta-domains}, a new concept designed to capture the key underlying features of datasets.
\textsc{Domain2Vec} maintains a vocabulary of meta-domains and uses a classifier to decompose any given dataset into a domain vector that corresponds to a distribution over this vocabulary.
These domain vectors enable the identification of the optimal data mixture for language model (LM) pretraining in a training-free manner under the \emph{\textbf{D}istribution \textbf{A}lignment \textbf{A}ssumption} (DA$^{2}$), which suggests that when the data distributions of the training set and the validation set are better aligned, a lower validation loss is achieved.
Moreover, \textsc{Domain2vec} can be seamlessly integrated into previous works to model the relationship between domain vectors and LM performance, greatly enhancing the efficiency and scalability of previous methods.
Extensive experiments demonstrate that \textsc{Domain2Vec} helps find the data mixture that enhances downstream task performance with minimal computational overhead.
Specifically, \textsc{Domain2Vec} achieves the same validation loss on Pile-CC using only $51.5\%$ of the computation required when training on the original mixture of The Pile dataset.
Under equivalent compute budget, \textsc{Domain2Vec} improves downstream performance by an average of $2.83\%$.
% \textsc{Domain2Vec} serves as a strong and efficient baseline for data mixture optimization in LM pretraining, offering insights into improving data efficiency in large-scale models.
\end{abstract}

\section{Introduction}
Through training on large-scale text corpora, Large Language Models (LLMs) have demonstrated remarkable generalization capabilities~\cite{touvron2023llama2openfoundation, openai2024gpt4technicalreport, yang2024qwen2technicalreport, deepseekai2024deepseekv2strongeconomicalefficient}.  The training datasets for LLMs are typically composed of multiple domains from diverse sources.
Recent research has shown that the mixture proportions of these domains referred to as the data mixture, can significantly influence the effectiveness of LMs~\cite{hoffmann2022an, xie2023data}, with data from one domain potentially affecting the performance on others~\cite{guo2022sample}.
Typically, the data mixtures used for training LLMs are determined heuristically or based on downstream performance metrics. 
However, these methods lack scalability and often yield suboptimal data mixtures.
Thus, identifying the optimal data mixture in a scalable and efficient manner remains a critical and challenging research question.
 
Recently, researchers have proposed various methods to determine the optimal data mixture. 
% In this paper, we categorize prior work into two main lines.
The first line of prior works implicitly adjusts the data mixture by selecting high-quality data from different domains or datasets~\cite{NEURIPS2024_3322a9a7,ankner2024perplexedperplexityperplexitybaseddata,thakkar2023selfinfluence}. 
% For instance,~\citet{NEURIPS2024_3322a9a7} propose using selective language models to choose tokens that align with the ideal data mixture. 
% Similarly,~\citet{ankner2024perplexedperplexityperplexitybaseddata} and~\citet{thakkar2023selfinfluence} filter out low-quality samples.
%filter out low-quality data at the sample level based on perplexity or influence scores.
The second line of work focuses on modeling the relationship between the data mixture and the performance of LLMs, and explicitly adjusts the data mixture across different datasets~\cite{rae2022scalinglanguagemodelsmethods, xie2023doremi, Sagawa*2020Distributionally, fan2023doge, ye2024datamixinglawsoptimizing, ge2024datamixingefficientbivariate, gu-etal-2024-cmr, NEURIPS2024_a4628e9f}. 
While prior work has shown promising results, there are some key issues:
\textbf{1) Computational Efficiency}: 
For example, although the proxy model in DoReMi~\cite{xie2023doremi} has only $280$M parameters, its estimated FLOPs are high to $3.7\times10 ^{19}$ for calculating on only 22 datasets. 
Moreover, The computational complexity of these methods will grow non-linearly as the number of datasets increases.
% Similarly, RHO-1~\cite{NEURIPS2024_3322a9a7} only calculates loss on specific tokens but still requires the entire sentence to be input into the model, leading to high computational costs.
\textbf{2) Lack of Scalability}:  
After establishing the functional relationship between data mixtures and model performance~\cite{ye2024datamixinglawsoptimizing, liu2024regmixdatamixtureregression}, if the dataset composition changes (e.g., by adding new datasets or filtering low-quality data, etc), previously fitted functions cannot be directly applied. 
This requires resampling new data mixtures, retraining proxy models, and refitting the functions, severely limiting the scalability of these methods.

To address these issues, we introduce \textsc{Domain2Vec}, a novel framework designed to vectorize datasets. 
This enables us to perform all operations for computing optimal mixing ratios in domain vector space, which has broad applicability when datasets change.
%, which maintains broad applicability without requiring model retraining when datasets change.
Specifically, \textsc{Domain2Vec} maintains a vocabulary of meta-domains, and we hypothesize that \emph{any dataset can be approximated as a linear combination of several meta-domains with a specific distribution}. 
This distribution could serve as the vector representation (domain vector) of a given dataset.

To efficiently identify the meta-domain composition of any given dataset, we propose to use a meta-domain classifier to generate the corresponding domain vector. 
Building upon \textsc{Domain2Vec}, we introduce the \emph{\textbf{D}istribution \textbf{A}lignment \textbf{A}ssumption} (DA$^{2}$) to find optimal data mixtures for LM pretraining. 
DA$^{2}$ states that \emph{lower validation loss can be achieved when the domain vector of the training dataset better aligns with the domain vector of the validation dataset}. 
Based on DA$^2$, we can easily find the optimal data mixture without training.

Moreover, \textsc{Domain2Vec} can be seamlessly integrated into prior works like RegMix~\cite{liu2024regmixdatamixtureregression}. 
Unlike previous methods that model the relationship between data mixtures and language model performance~\cite{liu2024regmixdatamixtureregression, ye2024datamixinglawsoptimizing}, 
we model the relationship between domain vectors provided by \textsc{Domain2Vec} and model performance, further enhancing efficiency and scalability of previous works.

In summary, we highlight our contributions as follows:
\begin{enumerate}
\item We introduce \textsc{Domain2Vec} to vectorize datasets and propose viewing datasets as combinations of meta-domains. We present an efficient pipeline for vectorizing datasets using a meta-domain classifier.
\item We propose the \emph{\textbf{D}istribution \textbf{A}lignment \textbf{A}ssumption} (DA$^{2}$), a training-free method for identifying the optimal data mixture. We further demonstrate how \textsc{Domain2Vec} can be seamlessly integrated into prior work to improve efficiency and scalability.
\item We validate the effectiveness of \textsc{Domain2Vec}+DA$^2$ and +RegMix in text generation and downstream tasks. Experimental results show that our method can accurately predict the performance of various data mixtures without training proxy models. Moreover, we can identify data mixtures that achieve downstream performance comparable to DoReMi~\cite{xie2023doremi}, using only $0.26$\% of its computational cost.
\end{enumerate}

\section{Domain2Vec}
\label{sec:domain2vec}
In this section, we introduce \textsc{Domain2Vec}, an algorithm that decomposes a dataset into a linear combination of various meta-domains and allows us to represent the underlying features of a dataset through a normalized vector.
We also outline a pipeline for constructing the vocabulary of \textsc{Domain2Vec} and training a meta-domain classifier.

\paragraph{Key Assumption.} 
\textsc{Domain2Vec} maintains a vocabulary, a set of meta-domains. 
Assume we have $n$ meta-domains $\mathcal D_j^*$ ($1\leq j \leq n)$, where $\mathcal D_j^*$ is represented as $\bm{e}_j$, a one-hot vector where only the $j$-th element is $1$.
We hypothesize that, for any given dataset $\mathcal D$, it could be represented as a domain vector $\bm{v}$ through a linear combination of these meta-domains. Specifically,
\begin{equation}
\label{eq:meta-domain-assumption}
\begin{aligned}
    \bm{v} \approx \sum_{j=1}^{n} v_j \cdot \bm{e}_j,
\end{aligned}
\end{equation}
where each element $v_j$ of $\bm{v}$ represents the projection (weight) of the dataset $\mathcal D$ on $\mathcal{D}_j^*$. 
Thus, $\bm{v}=[v_1, v_2, ...,v_{n}]^{\top}$ can be a representation (distribution) of the dataset $\mathcal{D}$ over the meta-domains.
However, an ideal approach for constructing these meta-domains remains to be established.
Next, we will introduce how we construct meta-domains from large-scale unlabeled text corpora.
\begin{figure*}[!t]
\centering
    \includegraphics[width=1.0\textwidth]{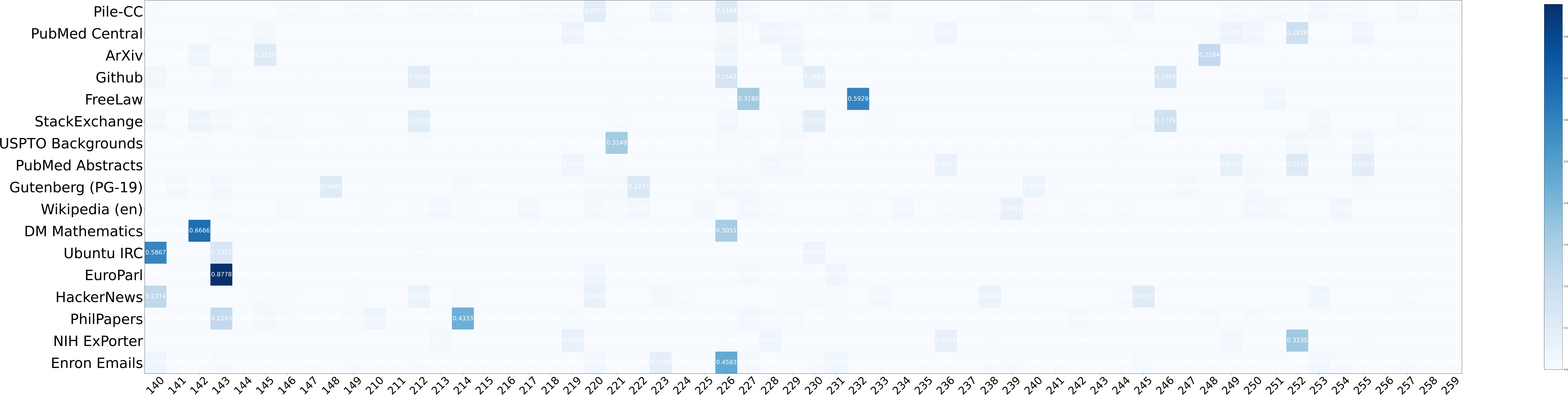 }
    \vspace{-1.0em}
    \caption{The domain vector of each sub-dataset of The Pile~\cite{DBLP:journals/corr/abs-2101-00027}, where each row corresponds to a sub-dataset and each column corresponds to a meta-domain. The higher the proportion of data belonging to a particular meta-domain, the closer the color of the corresponding cell is to \textcolor[RGB]{0,0,255}{blue}. We display distribution on some English meta-domains for clarity. The full picture is shown in Figure~\ref{fig:pile_en_full}.}
    \label{fig:pile_en}
    \vspace{-5pt}
\end{figure*}

\paragraph{Constructing the Vocabulary of \textsc{Domain2Vec}.} 
With the above key assumption, we define meta-domains as a collection of actual datasets that serve as a \emph{basis} in the domain vector space, allowing for linear combinations of these concrete datasets to represent any unknown domain in this space.
These constructed meta-domains, which could represent datasets from any source, should satisfy the following three properties, similar to the properties of a basis in linear algebra:

\begin{enumerate}
    \item \textbf{Spanning Set.} The domains that compose meta-domains should be as diverse and comprehensive as possible.
    \item \textbf{Linear Independence.} There should be distinct differences between these constructed meta-domains.
    \item \textbf{Computational Efficiency (Optional).} The method for constructing meta-domains should be computationally efficient.
\end{enumerate}

For diverse and comprehensive meta-domains, we collect data from more than $100$ coarse sources in English, Chinese\footnote{In this paper, we primarily aim at languages of English and Chinese.}, and Code.
After deduplication, we obtain around \textbf{$5.2$ TB} text data including more than $1$ billion documents.
The large corpora have a similar source composition as the standard large-scale LLM pretraining, including common crawl (CC), Wikipedia, social media platform, arXiv, code, news, books, etc. 
One could assume that the corpora already include as diverse and comprehensive contents as possible, corresponding to the requirement ``spanning set''~\footnote{Due to deduplication pre-processing and the native difference among the corpora, the requirement ``linear independence'' is also naturally satisfied.}.

\begin{figure}[!ht]
\vspace{-2mm}
\begin{center}
\centerline{ \includegraphics[width=0.8\columnwidth]{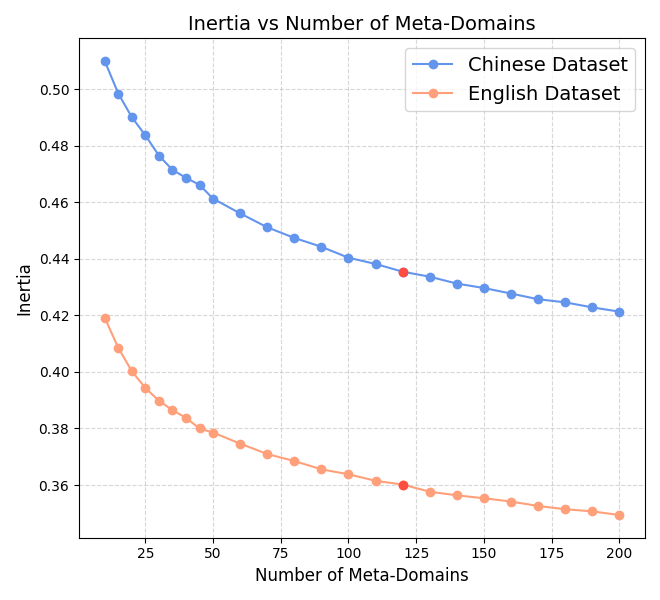}}
\vspace{-1.0em}
\caption{The number of meta-domains vs. Inertia.}
\label{fig:number_of_meta_domains}
\end{center}
\vspace{-8mm}
\end{figure}

After getting the corpora, we aim to extract the meta-domains in the corpora, that is, to divide the corpora into some (semantically) different clusters, to serve as the meta-domains. 
We employ $k$-means~\citep{macqueen1967some, arthur2006k} clustering algorithm to implement the separation and utilize \texttt{bge-small-en-v1.5} and \texttt{bge-small-zh-v1.5}~\citep{bge_embedding} to compute embeddings for the English and Chinese documents, respectively. 
See Figure~\ref{fig:number_of_meta_domains} for the relationship between the number of meta-domains and Inertia (measuring the distance between each data point and its centroid).
Besides, we divide the code data directly according to the programming language.
Ultimately, we construct $260$ ($120$ Chinese + $120$ English + $20$ Code) unique meta-domains. Each document in the corpora is labeled which meta-domain it originates from.

\paragraph{Meta-Domain Classifier.}
We now present our approach for representing an unseen dataset using the previously established meta-domains. 
The methodology is straightforward yet effective: we assign each document in the unseen dataset to its corresponding meta-domains and then calculate the aggregate distribution across all documents.
This comprehensive representation captures the overall domain characteristics of the entire dataset.
Formally, assume that there is a meta-domain classifier, for any given document~$doc \in \mathcal{D}$,
\begin{equation}
\begin{aligned}
\bm{p} = [p_1, p_2, p_3, ..., p_{n}]^\top = \mathrm{Classifier}(doc),
\end{aligned}
\label{eq:Meta-Domain Classifier}
\end{equation}
where $p_i$ represents the probability that $doc$ belongs to the $i$-th meta-domain such that $\lVert \bm{p} \rVert_1 = 1$.
For the unseen dataset $\mathcal{D}$, we can sample $N$ documents\footnote{In this paper, we set $N=1000$, which is enough for an accurate and stable domain vector.} then take the average of domain vectors of these samples.
Formally, the domain vector $\bm{v}$ of dataset $\mathcal{D}$ is,
\begin{equation}
\label{eq:distribution}
\begin{aligned}
\bm{v} \approx \frac{1}{N}\sum_{i=1}^{N} \bm{p}_i,
\end{aligned}
\end{equation}

Then, we could use the vector $\bm{v}$ to approximately represent the feature of any unseen dataset $\mathcal{D}$. 
Meanwhile, during the pretraining phase of LLMs, we typically have training datasets from many sources $\mathcal D_{train} = \{ \mathcal D_1, \mathcal D_2,...,\mathcal D_m \}$. 
We can convert each of these datasets into domain vectors following Equation~\ref{eq:Meta-Domain Classifier} and \ref{eq:distribution}.
Therefore, $\mathcal D_{train}$ can be approximately represented as $\boldsymbol V_{train} = [\boldsymbol{v}_1, \boldsymbol{v}_2,...,\boldsymbol{v}_m]$, where $\boldsymbol V_{train} \in \mathbb R^{n \times m}$ and $n$ is the number of meta-domains.

Specifically, we train a 260-class classifier to determine which meta-domain a given document originates from.
We finetune a \texttt{Qwen2-1.5b-base}~\citep{yang2024qwen2technicalreport} to balance accuracy and efficiency.
After training, the meta-domain classifier achieves a classification accuracy of $74.73\%$ on our constructed test set.
For further evaluating the performance of the meta-domain classifier, 
we also sample $1,000$ examples from each sub-dataset of The Pile~\citep{DBLP:journals/corr/abs-2101-00027}. 
Following Equation~\ref{eq:distribution}, we obtain domain vectors predicted by the meta-domain classifier for each sub-dataset, as shown in Figure~\ref{fig:pile_en}.
The distributions of sub-datasets of The Pile over meta-domains exhibit distinctive patterns.
This phenomenon indicates not only that the various meta-domains have significant semantic differences, but also that our classifier can accurately distinguish semantic features from different unseen datasets.

\section{Methodology}
\label{sec: section 3}
In this section, we first introduce the task formulation of the optimal data mixture discovery.
We then present methodologies for identifying the optimal data mixture using \textsc{Domain2Vec} without requiring additional training.
We introduce two approaches: the first is grounded in the \emph{\textbf{D}istribution \textbf{A}lignment \textbf{A}ssumption} (\textbf{DA$^2$}). 
Moreover, we demonstrate how our \textsc{Domain2Vec} can be integrated with previous works that model the relationship between mixture ratios and final performance, significantly enhancing the scalability of these existing approaches.

\subsection{Task Formulation}
\label{sec: task_form}
During the pretraining phase of LLMs, we typically collect training datasets $\mathcal D_{train} = \{ \mathcal D_1, \mathcal D_2,...,\mathcal D_m \}$ from $m$ sources (e.g., arXiv, Wikipedia, etc.). 
We also pre-define a validation set $\mathcal D_{valid}$, which is high-quality and indicative of final performance.
Note that $\mathcal D_{valid}$ is often independently and identically distributed with $\mathcal D_{train}$.
For example,~\citet{liu2024regmixdatamixtureregression} adopts Pile-CC~\citep{DBLP:journals/corr/abs-2101-00027} as the validation set and~\citet{gu2024data} adopts LIMA~\citep{zhou2023lima} as the validation set.
Accordingly, the data mixture $\boldsymbol{r} = [r_1, r_2, ..., r_m]^\top$, where $0 \leq r_i \leq 1$ and $\sum_{i=1}^m r_i = 1$, specifies the mixture ratio of the $m$ datasets. 
Let the trained LM be denoted as $\theta$, and the validation loss of the LM be denoted as $\mathcal{L}_\theta$. 
The objective of finding the optimal data mixture $\bm{r}^*$ is usually to minimize the validation loss, as shown formally in Equation~\ref{eq:optimized target}.
We denote $\mathcal L^{\mathcal D_{valid}}(\bm{r})$ as the validation loss of a LM pretrained on the data mixture $\bm{r}$.
\begin{equation}
\begin{aligned}
\label{eq:optimized target}
\boldsymbol{r}^* =  \arg\min_{\boldsymbol{r}} (\min_{\theta} \mathcal L_{\theta}^{\mathcal D_{valid}}(\boldsymbol{r})) \triangleq \arg\min_{\boldsymbol{r}} \mathcal L^{\mathcal D_{valid}}(\boldsymbol{r})
\end{aligned}
\end{equation}

\begin{figure*}[t]
\centering
    \includegraphics[width=1.0\textwidth]{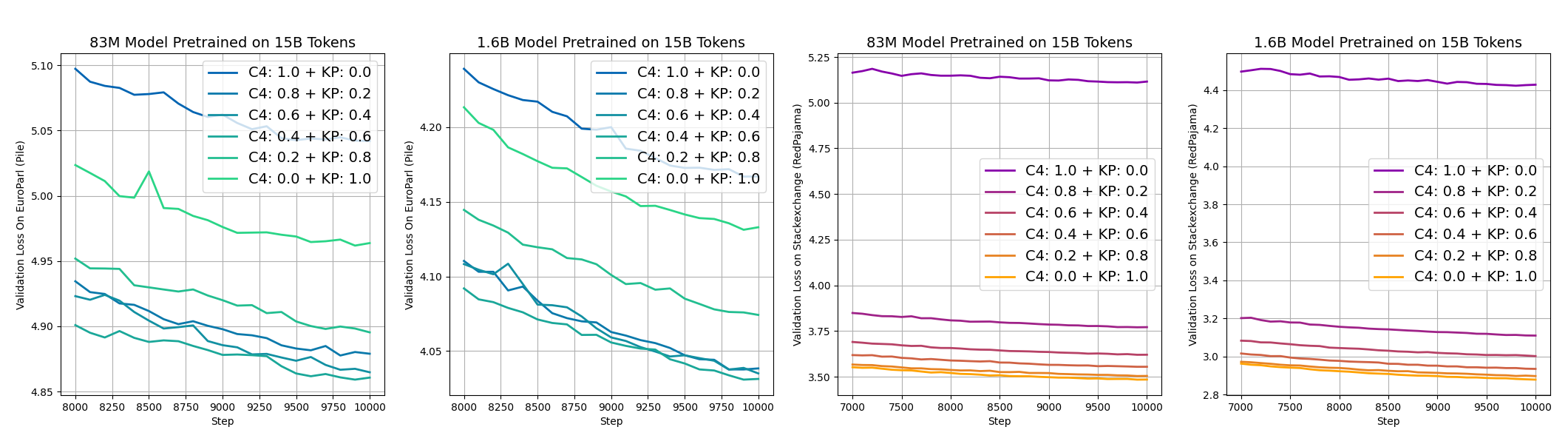}
    \vspace{-1em}
    \caption{The validation loss on the EuroParl (The Pile) and Stackexchange (RedPajama) of models trained using data mixture in Table~\ref{table: ood data mixture}. The validation loss on other validation sets are shown in Appendix~\ref{appendix: additional results}.}
    \label{fig:pilot study}
\end{figure*}

\subsection{Pilot Study: Mixture Ratio Ranking Holds across Model Sizes}
We first conduct a pilot study for a critical research question: \emph{Could the optimal data mixture generalize across different model sizes?}
If the answer is \emph{Yes}, it opens up the promising possibility that we could determine the optimal mixture ratio by simply training a small proxy model—or even more efficiently, without training any model at all.
To answer the questions, we mix C4~\citep{10.5555/3455716.3455856} and Knowledge Pile~\citep{fei2024queryccunearthinglarge} with different data mixtures ($0, 0.2, \cdots 1.0$) in Table~\ref{table: ood data mixture}.
We pretrain two LMs with $83$M and $1.6$B parameters from scratch using the standard LM loss.
During pretraining, we evaluate the validation loss of models trained with different mixture ratios on $20$ subsets of The Pile~\citep{DBLP:journals/corr/abs-2101-00027} and RedPajama~\citep{together2023redpajama}, as shown in Figure~\ref{fig:pilot study}.
The results of more validation sets can be seen in Figures~\ref{fig:pilot study 1} and~\ref{fig:pilot study 2}.
There are two findings:
%\vspace{-2mm}
\begin{itemize}
    \item \emph{An optimal mixture ratio exists for each validation set, with rankings varying significantly across different validation sets.}
    %\vspace{-2mm}
    \item \emph{For the same validation set, the ranking of data mixture ratios remains nearly unchanged with varying model sizes.} We calculate the correlation coefficients of data mixture rankings between the $83$M model and the $1.6$B model across diverse validation sets. The analysis yields a Spearman coefficient of $0.9743$ and a Pearson coefficient of $0.9947$, providing robust statistical evidence for this consistency. These exceptionally high correlation values strongly support our finding that optimal mixture ratios are largely invariant to model size when evaluated on the same validation benchmark.
\end{itemize}
%\vspace{-2mm}
These finding aligns with prior work by~\citet{liu2024regmixdatamixtureregression}, which indicates that it is possible to find the optimal data mixture without training (Section ~\ref{sec: da2}) or simply training small models (Section ~\ref{sec: domain2vec-func}).

\subsection{\textbf{D}istribution \textbf{A}lignment \textbf{A}ssumption~(DA$^2$)}
\label{sec: da2}
We introduce how we directly apply our proposed \textsc{Domain2Vec} on finding optimal data mixture.
We notice an intuitive principle that a lower validation loss $\mathcal{L}^{\mathcal D_{valid}}$ is achieved when the data distribution of the training set is better aligned with the given validation set\footnote{We provide the detailed description in the Appendix~\ref{appendix: Description}.}.
One essential question is that \emph{How do we model the data distribution of various datasets?}
Fortunately, according to Section~\ref{sec:domain2vec}, for the training dataset $\mathcal D_{train}$, we obtain the vector representation $\bm V_{train} \in \mathbb R^{n \times m}$, which models semantic features of $\mathcal D_{train}$.
Correspondingly, for the validation set $\mathcal D_{valid} $, we also have its vector representation $\bm{v}_{valid}$.
After mixing $\mathcal D_{train}$ with a data mixture $\bm{r}$, the final distribution over meta-domains of $\mathcal D_{train}$ is $\bm{v}_{train} = \bm V_{train}\cdot \bm{r}$. 
Therefore, based on the distribution alignment assumption, Equation \ref{eq:optimized target} can be equivalently written as:
\begin{equation}
\label{eq:distribution consistency assumption}
\begin{aligned}
\boldsymbol{r}^* =  \arg\min_{\boldsymbol{r}} \mathrm{Dist}(\boldsymbol V_{train}\cdot \boldsymbol{r}, \boldsymbol{v}_{valid})
\end{aligned}
\end{equation}
where $\mathrm{Dist}(\cdot, \cdot)$ is a distance function used to measure the similarity between two vectors. 
Theoretically, numerous distance function options are available, including Wasserstein (optimal transport) distance, Euclidean distance, etc.
In this paper, we use Huber Loss~\cite{10.1214/aoms/1177703732, hastie2009elements} between two vectors to measure the distance.
We also discuss the choice of different distance functions in Appendix~\ref{appdenix: Different Distributional Measures}. We present the pseudo code of \textsc{Domain2Vec}+DA$^2$ in Appendix~\ref{appendix: Algorithm}.

\subsection{Applying Domain2Vec to Prior Work}
\label{sec: domain2vec-func}
There is one typical line of research focused on determining optimal mixture ratios, which aims to model the relationship between these ratios and the final validation loss using various functional approaches. 
That is,  these approaches model $\mathcal L^{\mathcal D_{valid}}(\bm{r}) = f(\bm{r})$ where $f(\cdot)$ can take various reasonable forms as proposed in previous works. For example,
\begin{itemize}
    \item Data Mixing Law~\citep{ye2024datamixinglawsoptimizing} adopts $f(\bm{r}) = c_i + k_i \cdot \exp \left( \sum_{j=1}^m t_{ij} \cdot r_j \right)$ to predict the validation loss on training domain $i$, where $c_i, k_i, t_{ij}$ are all undetermined parameters to fit.
    \item RegMix~\citep{liu2024regmixdatamixtureregression} initially adopts a Linear Regression approach, modeling the validation loss as $f(\bm{r}) = \bm{w}^\top \bm{r}$ where $\bm{w}$ needs fitting. Furthermore, it advances this concept by employing LightGBM~\citep{ke2017lightgbm} to more effectively fit the function $f(\cdot)$.
\end{itemize}

We can directly integrate \textsc{Domain2Vec} with these approaches without modifying their core function, but instead perform the computations in the domain vector space. 
Thereby, we address two inherent limitations of these approaches: (1) \textbf{Efficiency}: for modeling $f(\cdot)$ with $m$ variables $r_{1\cdots m}$ \footnote{$m$ can scale to to $10^4$ in modern LLM training. 
For example, Fineweb~\citep{penedo2024the} consists of over $30$k data dumps.}, it is expected to run experiments $O(m^2)$ times for different $\bm{r}$ to collect fitting points; (2) \textbf{Scalability}: When a new training source is introduced, one must re-collect fitting points and re-fit $f(\cdot)$, which lacks of scalability.

Specifically, we novelly build the relationship $f_i(\cdot)$ between the validation loss on the $i$-th meta-domain $\mathcal D_{i}^{*}$
(notated as $\mathcal L^{D_{i}^*}$)
and the domain vector $\bm{v}_{train}$ after mixing training datasets by ratio $\bm{r}$, that is,  $\bm V_{train}\cdot\bm{r}$. Formally, for each meta-domain, we have
\begin{equation}
\label{eq:linear regression model}
\mathcal L^{\mathcal D_{i}^*} (\bm{r})= f_i(\bm{v}_{train}) = f_i(\bm V_{train}\cdot\bm{r}), 1\leq i \leq n.
\end{equation}
Equation~\ref{eq:linear regression model} enables the prediction of validation loss on any meta-domain given a data mixture, which is also the function that we aim to fit.
For unseen validation dataset, recall that any dataset including $\mathcal D_{valid} $ can also be viewed as a linear addition of meta-domains and the domain vector of $\mathcal D_{valid}$ is denoted as $\bm{v}_{valid} = [q_1, q_2, \cdots, q_{n}]^\top$. 
Therefore, we have
\begin{equation}
\label{eq:addictive}
\begin{aligned}
\mathcal L^{\mathcal D_{valid}} (\bm{r}) &= \sum \limits_{i=1}^{n} q_i \cdot \mathcal  L^{\mathcal D^{*}_{i}} (\bm{r}) =  \sum \limits_{i=1}^{n} q_i \cdot f_i(\bm{v}_{train}) \\ &=  \sum \limits_{i=1}^{n} q_i \cdot f_i(\bm V_{train}\cdot\bm{r}).
\end{aligned}
\end{equation}
Now, we connect validation loss to the mixture ratio in the the domain vector space via our proposed \textsc{Domain2Vec}.
It is feasible to search the optimal mixture ratio $r^\star$ by minimizing $\mathcal L^{\mathcal D_{valid}} (\bm{r})$.
Note that this connection is built only on the top of the meta-domains (i.e., $f_i$ for $1\leq i \leq n$), and can adapt with no cost to (1) any unseen validation set; 
(2) any unseen training set; (3) any number of training sets.
Thanks to this property, we realize the efficiency and scalability by \textsc{Domain2Vec} for prior approaches.

\begin{figure}[tbp]
  \centering
  %\vspace{-0.75em}
  \includegraphics[width=0.825\linewidth]{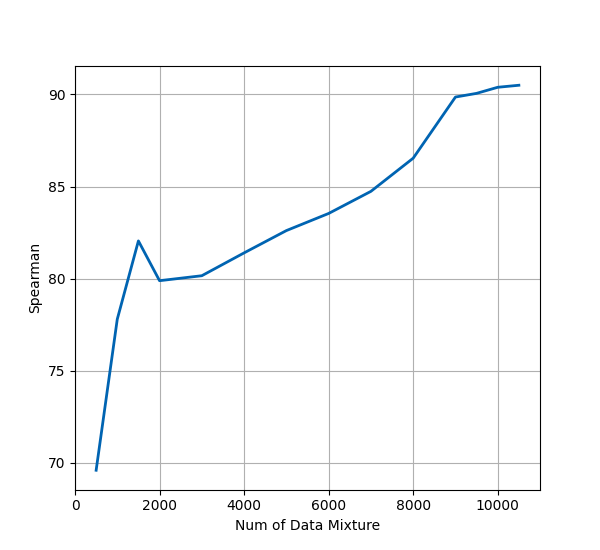}
  %\vspace{-1.5em}
  \caption{Relationship between the number of trained data mixtures and the Spearman correlation.}
  \label{fig:lightGBM}
  \vspace{-10pt}
\end{figure}

Following RegMix~\citep{liu2024regmixdatamixtureregression},
we use LightGBM~\citep{ke2017lightgbm} as $f(\cdot)$ to fit Equation~\ref{eq:linear regression model} for each meta-domain (named as \textsc{Domain2Vec}+RegMix).
The pseudo code of \textsc{Domain2Vec} + RegMix are shown in Appendix~\ref{appendix: Algorithm}.
We sample $10,500$ diverse mixture ratios from a Dirichlet distribution  and we get the validation losses on each meta-domains by training $10,500$ small LMs.
We also reserve some mixture ratios as testset and run experiments for evaluating whether fitted function $f(\cdot)$ can accurately predict the validation loss for unseen mixture ratios.
For various mixture ratios in the testset, we use the Spearman coefficient to measure the correlation between the predicted ranking and the actual ranking of performance under unseen mixture ratios. Note that we adopt correlation coefficient because it is a more general metric than mean loss error with the goal to find the better mixture ratio than others.
Moreover, the pilot study suggests that the predicted ranking holds across model sizes while the predicted loss becomes meaningless for inconsistent model sizes.
As shown in Figure~\ref{fig:lightGBM}, the Spearman coefficient increases with the number of mixture ratios that we use for training and collecting fitting points.
And finally we get an over $90\%$ Spearman coefficient, which is quite accurate for predicting a good mixture ratio for various meta-domains.

\section{Experiments}
In this section, we describe the implementation and results to demonstrate how \textsc{Domain2Vec} identifies the optimal data mixture with reduced computational cost.
The goal of optimizing the data mixture is to \emph{enhance the performance of LMs}. 
The performance of LMs can be evaluated from two perspectives:
1) Text generation, often measured by LM loss on a validation set. We aim to minimize the validation loss through finding the optimal data mixture; 
2) Downstream task performance. 
The objective is to optimize task performance.
As an overview for exprimental results, by applying \textsc{Domain2Vec}, we can accurately predict the ranking of data mixtures under various settings (e.g., training and validation sets).
We also achieve a validation loss comparable to that of the original data mixture from The Pile while using only 51.5\% of the computational resources.
Moreover, we use only $0.26\%$ of the computational costs required by DoReMi to find a data mixture with performance comparable to strong baselines like DoReMi.

\subsection{Validation Loss Minimization}
\label{sec:task1}

% \begin{wraptable}{R}{4.75cm}
% %\vspace{-7mm}
% \caption{The data mixture we used to mix C4~\cite{10.5555/3455716.3455856} and Knowledge Pile~\cite{fei2024queryccunearthinglarge}.}
% \label{table: ood data mixture}
% \begin{center}
% \resizebox{1.0\linewidth}{!}{
% \begin{tabular}{l|cccccc}
% \toprule
% \multicolumn{1}{l}{\textbf{Dataset} }  & \multicolumn{6}{c}{\textbf{Data Mixture} } \\                          
% \midrule
% C4  & 0 & 0.2 & 0.4 & 0.6 & 0.8 & 1.0 \\
% \midrule
% Knowledge Pile & 1.0 & 0.8 & 0.6 & 0.4 & 0.2 & 0.0 \\
% \bottomrule
% \end{tabular}
% }
% \end{center}
% %\vspace{-2.5mm}
% \end{wraptable}

\begin{figure*}[tbp]
\centering
    \includegraphics[width=1.0\textwidth]{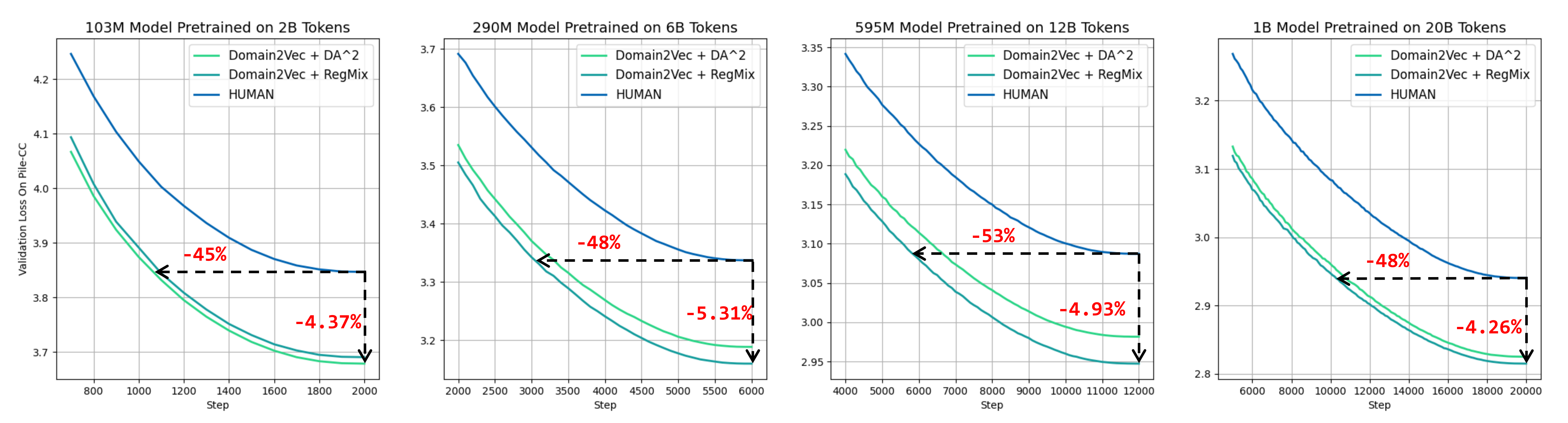}
    \vspace{-1em}
    \caption{The validation loss on the Pile-CC subset. \textsc{Domain2Vec} achieves the comparable validation loss of Human (The model using original data mixture from The Pile), but uses only $51.5\%$ training computational costs of Human. Using the same training cost, \textsc{Domain2Vec} can reduce the validation loss by approximately $4.72\%$ compared to Human.}
    \label{fig:pile_loss}
    \vspace{-5pt}
\end{figure*}

\paragraph{Dataset \& Data Mixture.}
We design some training and validation datasets to evaluate the performance to minimize the validation loss of our methods. Our training datasets include C4~\cite{10.5555/3455716.3455856} and Knowledge Pile~\cite{fei2024queryccunearthinglarge}.
C4 is a colossal and cleaned version of Common Crawl corpus.
Knowledge Pile is a high-quality dataset that significantly improves the performance of LLMs in knowledge-related and mathematical reasoning tasks.
We conduct our experiments on various validation datasets to perform comprehensive evaluation.
We select $20$ validation datasets from The Pile~\cite{DBLP:journals/corr/abs-2101-00027} and RedPajama~\cite{together2023redpajama}.
Since the optimal mixture ratio varies among the validation datasets, we instead predict the performance ranking across different preset mixture ratios.
Specifically, we mix C4 and Knowledge Pile with different data mixtures as the training set as shown in Table~\ref{table: ood data mixture}. 
%\vspace{-4mm}
\begin{table}[h]
\caption{The preset data mixture ratios.}
\label{table: ood data mixture}
% %\vspace{-1mm}
\begin{center}
\resizebox{0.75\linewidth}{!}{
\begin{tabular}{l|cccccc}
\toprule
\multicolumn{1}{l}{\textbf{Dataset} }  & \multicolumn{6}{c}{\textbf{Data Mixture} } \\   
\midrule
C4  & 0 & 0.2 & 0.4 & 0.6 & 0.8 & 1.0 \\
\midrule
Knowledge Pile & 1.0 & 0.8 & 0.6 & 0.4 & 0.2 & 0.0 \\
\bottomrule
\end{tabular}
}
\end{center}
\end{table}
\vspace{-6mm}

\paragraph{Training \& Evaluation Setup.}
We pretrain LLaMA-like~\cite{dubey2024llama3herdmodels} models with $83$M and $1.6$B parameters from scratch using standard language modeling loss. 
Both models have a batch size of $1.5$M tokens and a maximum sequence length of $4,096$.
We use the AdamW optimizer~\cite{loshchilov2017decoupled} with gradient clipping at $1.0$.
The learning rate linearly warms up to 2e-4 over the first $100$ steps, then decays to 2e-5 using a cosine scheduler over $10,000$ steps. 
More parameters are detailed in Table~\ref{table: model parameters}.
Then, we evaluate \textsc{Domain2Vec} using the Spearman and Pearson correlation coefficient between the predicted ranking and the actual ranking. 
We compare \textsc{Domain2Vec} with randomly ranking and an embedding-based baseline, denoted as $k$NN.
Specifically, we use \texttt{bge-small-v1.5} to obtain embeddings and apply mean pooling to generate unique embeddings for each dataset and meta-domain.
We then apply $k$NN based on Euclidean distance to compute the probability distributions of training and test datasets originating from each meta-domain, treating these distributions as new domain vectors.

\paragraph{Experimental Results.}
We present the validation loss curves for various data mixtures in Figure~\ref{fig:pilot study} and Appendix~\ref{appendix: additional results}. 
It can be observed that, on most validation sets, incorporating data from Knowledge Pile significantly reduces validation loss.
% This indicates the high quality of the training data in the Knowledge Pile.
% Then, we sample $10,000$ samples from C4 and Knowledge Pile respectively, and $1,000$ samples from each validation set.
We apply two \textsc{Domain2Vec}-based methods described in Section~\ref{sec: section 3} to rank the data mixture from Table~\ref{table: ood data mixture}.

As demonstrated in Table~\ref{table: results of c4 and kp}, the ranking predicted by \textsc{Domain2Vec} exhibits a strong positive correlation with the actual ranking, significantly outperforming random guessing and $k$NN.
The effectiveness of the $k$NN method partially validates the rationale behind our meta-domain vocabulary construction.
% Interestingly, we find that \textsc{Domain2Vec} + RegMix even predict that a mixture of $20\%$ Knowledge Pile and $80\%$ C4 could achieve the lowest validation loss on C4 validation set from RedPajama.
% We hypothesize that this is due to the higher data quality of Knowledge Pile compared to C4, as well as the overlap between these two datasets in certain meta-domains. 
% As a result, incorporating a portion of Knowledge Pile into the mixture likely enhances the training of C4.
It is also important to note that our method is a \emph{training-free approach}, unlike prior works that rely on training small proxy models to rank data mixtures. 
Despite the  more challenging setup, our method accurately predicts the rankings of different data mixtures. 
\begin{table}[h]
%\vspace{-4mm}
\caption{The results of deploying the \textsc{Domain2Vec} to predict the ranking of different Validation sets.}
\label{table: results of c4 and kp}
%\vspace{-1mm}
\begin{center}
\resizebox{1.0\linewidth}{!}{
\begin{tabular}{l|ccccc}
\toprule
\multicolumn{1}{l}{\textbf{Metrics} }  & \textbf{Random} & \textbf{$k$NN}& \textbf{\textsc{Domain2Vec}+DA$^2$} & \textbf{\textsc{Domain2Vec}+RegMix}\\                \midrule
Pearson  & 0.0300 & 0.4014 & \textbf{0.5833} & 0.3881 \\
\midrule
Spearman & 0.0497 & 0.3543 & \textbf{0.6657} & 0.4629 \\
\bottomrule
\end{tabular}
}
\end{center}
\vspace{-10pt}
\end{table}

\subsection{Downstream Task Performance Maximization}
\label{sec:task2}
In this section, we demonstrate how \textsc{Domain2Vec} can be used to identify the optimal data mixture for maximizing downstream task performance.
One challenge is modeling the relationship between data mixture and downstream performance.
Fortunately,~\citet{liu2024regmixdatamixtureregression} finds that \emph{validation loss on Pile-CC correlates most strongly with downstream performance across their evaluations}. 
To align with prior work, we follow and use the same validation datasets as \citet{liu2024regmixdatamixtureregression}. 
Thus, our proxy goal is to identify a data mixture that minimizes validation loss on Pile-CC. 
\emph{Experimental results show that~\textsc{Domain2Vec} predicts a data mixture with performance comparable to DoReMi~\cite{xie2023doremi}, while using only $0.26\%$ computational cost}.

\begin{table*}[!t]
%\vspace{-2mm}
\caption{\textbf{Average} downstream task performance of different models pretrained on different data mixtures. 
Similar to~\citet{liu2024regmixdatamixtureregression}, Human refers the original data mixture from The Pile. 
Pile-CC is a golden training set which can $100\%$ correspond to validation set to validate our propose $DA^2$.
All the data mixtures are shown in Table~\ref{table: data mixture of pile} and Table~\ref{table: our data mixture of pile}. 
The calculated data mixture are shown in Table~\ref{table: 106M models' main results}.}
\label{table: main results}
\begin{center}
\footnotesize
\resizebox{0.98\textwidth}{!}{
\begin{tabular}{l|cccccc}
\toprule
\multicolumn{1}{l}{\textbf{Benchmark} }  &
\multicolumn{1}{c}{\textbf{Human} }  &
\multicolumn{1}{c}{\textbf{DoReMi} }  &
\multicolumn{1}{c}{\textbf{Pile-CC Only} }  &
\multicolumn{1}{c}{\textbf{RegMix} }  &
\multicolumn{1}{c}{\textbf{\textsc{Domain2Vec} $+$ DA$^{2}$} }  &
\multicolumn{1}{c}{\textbf{\textsc{Domain2Vec} $+$ RegMix} } 
\\                          
\midrule
Social IQA  & 0.367 & 0.380 & 0.381 & 0.382 & 0.372 & 0.375 \\
HellaSwag   & 0.319 & 0.346 & 0.351 & 0.351 & 0.335 & 0.338 \\
PiQA        & 0.615 & 0.639 & 0.644 & 0.647 & 0.635 & 0.639 \\
OpenBookQA  & 0.264 & 0.275 & 0.276 & 0.276 & 0.275 & 0.272 \\
Lambada     & 0.199 & 0.240 & 0.247 & 0.241 & 0.219 & 0.232 \\
SciQ        & 0.710 & 0.695 & 0.688 & 0.708 & 0.701 & 0.701 \\
ARC Easy    & 0.411 & 0.428 & 0.436 & 0.438 & 0.427 & 0.426 \\
COPA        & 0.621 & 0.651 & 0.660 & 0.653 & 0.638 & 0.641 \\
RACE        & 0.274 & 0.291 & 0.288 & 0.288 & 0.279 & 0.282 \\
LogiQA      & 0.272 & 0.275 & 0.272 & 0.272 & 0.269 & 0.278 \\
WinoGrande  & 0.512 & 0.516 & 0.515 & 0.513 & 0.513 & 0.510 \\
MultiRC     & 0.521 & 0.528 & 0.515 & 0.529 & 0.524 & 0.534 \\
\rowcolor{gray!20}
Average Performance & 0.424 & 0.439 & 0.439 & 0.441 & 0.432 & 0.436\\
\midrule
\multirow{2}{*}{Estimated FLOPs}
 & \multirow{2}{*}{$0$}
 & $3.7 \times 10^{19} $ & \multirow{2}{*}{$0$} & $3.5 \times 10^{18}$ & $9.66 \times 10^{16}$  & $9.66 \times 10^{16}$ \\
 &  & $(100\%)$ &  &  $(9.46\%)$ &  $(0.26\%)$ & $(0.26\%)$ \\ 
\bottomrule
\end{tabular}
}
\end{center}
\end{table*}

% \subsection{Experimental setup}
\paragraph{Datasets \& Baselines.}
We follow RegMix~\citep{liu2024regmixdatamixtureregression} and use The Pile~\cite{DBLP:journals/corr/abs-2101-00027} as our training datasets. The Pile is an 825 GB English text corpus used for LLM pretraining. 
In line with RegMix, we use only the 17 components of The Pile that do not have copyright issues.
Our goal is to identify the data mixture that minimizes validation loss on the Pile-CC subset to improve downstream task performance.
We compare our approach with several baselines, including Human (the original data mixture), DoReMi~\cite{xie2023doremi}, and RegMix~\cite{liu2024regmixdatamixtureregression}. 
The Pile-CC Only baseline (which trains the model solely on the Pile-CC subset) is included to verify the strong correlation between Pile-CC validation loss and downstream performance.
The data mixtures for each baseline are shown in Table~\ref{table: data mixture of pile}.

\paragraph{Training \& Evaluation Setup.}
We pretrain LLaMA-like~\cite{dubey2024llama3herdmodels} models from scratch using standard language modeling loss with model sizes ranging from $106$M to $1$B parameters. Following~\citet{hoffmann2022trainingcomputeoptimallargelanguage}, the token count for each model is $20$ times corresponding parameter size. 
All models adopt a batch size of $1$M tokens and a maximum sequence length of $4,096$.
We apply the AdamW~\cite{loshchilov2017decoupled} optimizer with gradient clipping at $1.0$. 
The learning rate linearly warms up to 6e-4 over $1,000$ steps, then decays to $0$ using a cosine scheduler at the end of training.
More parameters are detailed in Table~\ref{table: model parameters}.
For evaluation, we track the performance on Pile-CC validation loss across different model sizes.
Besides, we evaluate the performance of different data mixture using following benchmarks: Social IQA~\cite{sap-etal-2019-social}, HellaSwag~\cite{Zellers2019HellaSwagCA}, PiQA~\cite{Bisk2019PIQARA}, OpenBookQA~\cite{Mihaylov2018CanAS}, Lambada~\cite{paperno-etal-2016-lambada}, SciQ~\cite{welbl-etal-2017-crowdsourcing}, ARC Easy~\cite{Clark2018ThinkYH}, COPA~\cite{gordon-etal-2012-semeval}, RACE~\cite{lai-etal-2017-race}, LogiQA~\cite{10.5555/3491440.3491941}, WinoGrande~\cite{10.1145/3474381}, and MultiRC~\cite{khashabi-etal-2018-looking}. 
We utilize LM Evaluation Harness~\cite{eval-harness} to evaluate these models and report the average score across 0-shot to 5-shot settings in Table~\ref{table: main results}.

\paragraph{Implementation Details.}
% First, We sample $1,000$ documents from each subset of The Pile and Pile-CC validation set, using the meta-domain Classifier to calculate the domain vector for each dataset. 
% Using these domain vectors, 
We predict the optimal data mixture by applying Equation~\ref{eq:distribution consistency assumption}~(\textsc{Domain2Vec}+DA$^{2}$) and Equation~\ref{eq:addictive}~(\textsc{Domain2Vec}+RegMix).
We generate $100,000$ data mixtures from a Dirichlet distribution based on the token distribution of these components.
Using these mixtures, we predict the optimal data mixture by our proposed two methods.
We select top-$100$ predicted data mixtures and average them as the final data mixture. This trick is aligned with previous work~\cite{liu2024regmixdatamixtureregression} for more accurate and stable results.
As a stardard practice, each subset of The Pile is trained for at most one epoch. 
When optimizing the mixture ratio $\bm{r}=[r_1, r_2, \cdots, r_m]^\top$, other than the common restrictions $0\leq r_{1\cdots m}\leq1$ and $\sum_{i=1}^m r_i = 1$, note that there is another data amount restriction, that is, $ \#TotalTokens \cdot r_i \leq |\mathcal D_i|$, which is to remove data mixtures which require exceeding tokens in some subsets. 
Therefore, the optimal data mixture predicted by \textsc{Domain2Vec} could vary depending on the number of trained tokens, as well as the size of the models. This restriction is different with Section~\ref{sec:task1} where each dataset size is seen as unlimited.

\vspace{-5pt}
\paragraph{Experimental Results.}
As shown in Figure~\ref{fig:pile_loss}, \emph{our proposed \textsc{Domain2Vec + DA$^{2}$} and \textsc{Domain2Vec + RegMix} significantly improve training efficiency on Pile-CC compared to Human}.
Specifically, \textsc{Domain2Vec + DA$^{2}$} and \textsc{Domain2Vec + RegMix} require only about $55.38\%$ and $51.50\%$ of the training steps, respectively, to achieve the same validation loss as Human.
Compared to Human under the same compute budget, \textsc{Domain2Vec + DA$^{2}$} and \textsc{Domain2Vec + RegMix} reduce validation loss by approximately 4.04\% and 4.64\%, and improves downstream performance by an average of $1.89\%$ and $2.83\%$, respectively.
In Table~\ref{table: main results}, we report the average performance of LMs trained on data mixtures from various baselines across a range of downstream tasks.
``Pile-CC only'' shows a 3.54\% average accuracy improvement over Human, indicating that training on more tokens from Pile-CC enhances downstream performance.
Importantly, ``Pile-CC only'' is good when we regard Pile-CC as validation set. 
However, in a more practical scenario where validation set is somewhat else, we cannot manually find such a golden training set which can $100\%$ correspond to validation set.
To this end, we can use our proposed Domain2Vec to get a comparable downstream performance with lowest cost by mixing datasets from different sources.
% This suggests that different datasets may mutually benefit training for each other. 
Notably, \emph{\textsc{Domain2Vec + DA$^{2}$} and \textsc{Domain2Vec + RegMix}, using only about 0.26\% of the FLOPs required by DoReMi, achieve performance comparable to DoReMi, RegMix}, which demonstrates the computational efficiency of \textsc{Domain2Vec}.

\begin{figure}[tbp]
%\vspace{-2mm}
\begin{center}
\centerline{\includegraphics[width=0.95\columnwidth]{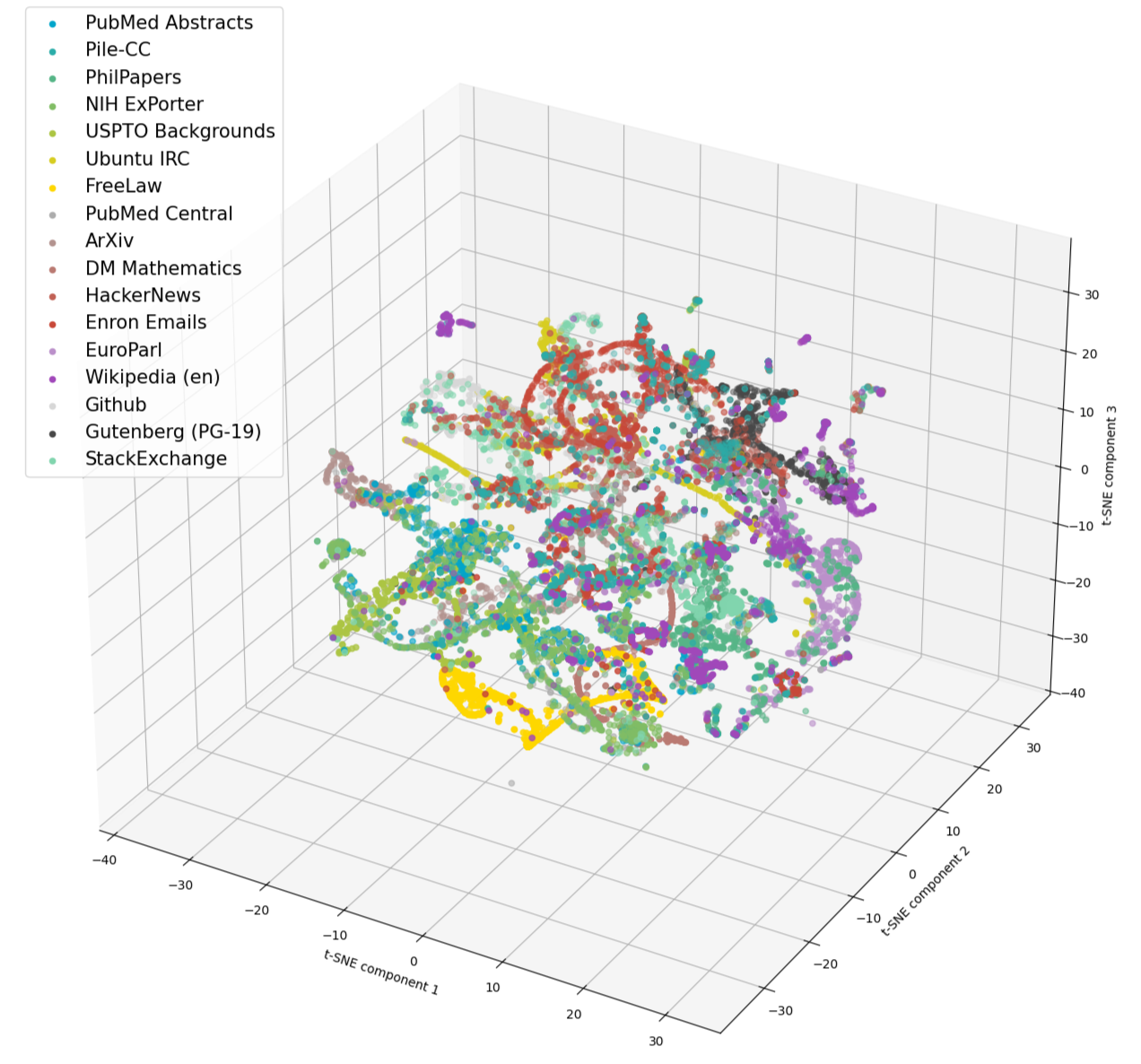}}
%\vspace{-4mm}
\caption{Visualization (t-SNE) of domain vectors of The Pile.}
\label{fig:t-sne-3d}
\end{center}
\vspace{-20pt}
\end{figure}

\paragraph{Visualization.}
To investigate further, we employ t-SNE~\cite{van2008visualizing} to visualize the domain vectors of each component in The Pile, as shown in Figure~\ref{fig:t-sne-3d}.
% The visualization reveals that different datasets can share data from the same meta-domain, with datasets like Pile-CC, Wikipedia, and PhilPapers encompassing data from multiple meta-domains. This overlap suggests that the domain vector effectively captures the underlying features of various datasets.
This visualization reveals several desirable properties of the learned vectors. The representation space exhibits strong clustering behavior where semantically related datasets naturally group together, indicating effective capture of domain-specific characteristics. Related domains such as academic literature (PubMed, arXiv) and technical repositories (GitHub, StackExchange) demonstrate spatial coherence, while maintaining well-defined yet flexible boundaries between different domains. The representation spans diverse domains in The Pile, demonstrating robust generalization capabilities across heterogeneous data types. 
% The balanced density distribution across clusters indicates stable embedding quality that should facilitate consistent performance in downstream applications.

% \vspace{-5pt}
\subsection{Discussion on Overfitting} 
We noticed that some readers interpret our approaches as a form of ``overfitting'': optimizing on a selected validation set. We offer the following explanations:

\begin{itemize}
\item The validation set that we define is actually a guide dataset, which is a necessary requirement for optimization data mixture and a common setting in related works (see Section~\ref{sec: task_form}). 
\item In Section~\ref{sec:task1}, we conduct experiments on various validation sets, and the performance demonstrates good stability. In fact, our proposed DA$^2$ does not even require training, thus "overfitting" is not applicable. 
\item In Section~\ref{sec:task2}, we choose Pile-CC as the validation set but ultimately test model performance on benchmarks from 12 downstream tasks, further preventing overfitting risks.
\end{itemize}

\section{Related Work}
Recent research on optimizing data mixture can be broadly divided into two lines.
The first line implicitly adjusts data mixture by down-sampling from various datasets based on data quality. 
For example,~\citet{NEURIPS2024_3322a9a7} propose RHO-1, which uses Selective Language Models to select tokens that align the data mixture with the ideal ratio. 
Instead of token-level selection, ~\citet{ankner2024perplexedperplexityperplexitybaseddata} filter low-quality samples using the perplexity of small reference models. ~\citet{thakkar2023selfinfluence} demonstrate that the Influence Score can guide data re-weighting, while their subsequent work introduces an online data selection method that eliminates the need for reference models.

The second line focuses on explicitly adjusting data mixture by modeling the relationship between data mixture and language model performance.
The simplest approach is to observe the performance of various data mixtures and select the optimal one, as done during Gopher training~\cite{rae2022scalinglanguagemodelsmethods}. 
This is costly and difficult to scale for larger models.
~\citet{xie2023doremi} propose DoReMi and use a small proxy model to re-weight data from different domains, improving training efficiency for larger models. 
However, DoReMi still requires a pre-trained reference model, adding computational costs and making it hard to define an ideal reference model.
% In response,~\citet{fan2023doge} introduce DoGE, which uses the min-max optimization to train a proxy model for obtaining better domain weights. This method assigns higher weights to domains that help learning in others or are more difficult to learn themselves.
% ~\citet{chen2023skillit} propose a skills-based framework to dynamically adjust data mixtures during model training. 
Some works aim to model the functional relationship between data mixture and the LM performance.
Inspired by scaling laws~\cite{kaplan2020scalinglawsneurallanguage, hoffmann2022trainingcomputeoptimallargelanguage}, ~\citet{ye2024datamixinglawsoptimizing} introduce Data Mixing Laws, which describe this relationship using an exponential form. 
~\citet{ge2024datamixingefficientbivariate} propose BiMix, a scaling law that considers both compute and data mixture.
~\citet{NEURIPS2024_a4628e9f} and~\citet{Wang2025learning} develop scaling laws for continual pretraining, and how mixture ratio as one variable impacts LM loss is modeled.
Recently, ~\citet{liu2024regmixdatamixtureregression} propose Linear Regression to model the validation loss across different data mixtures, showing a strong and promising performance.

All these prior works face two main issues:
\textbf{1) Computational Efficiency}:
For example, the estimated FLOPs for DoReMi and RegMix reach $3.7\times10 ^{19}$ and $3.5\times10 ^{18}$, respectively, when applied to approximately 20 datasets. Moreover, the computational complexity of these methods will grow non-linearly as the number of datasets increases.
\textbf{2) Lack of Scalability}:
When the components of the training dataset change (e.g., adding some new datasets), previous methods~\cite{ye2024datamixinglawsoptimizing, liu2024regmixdatamixtureregression} require resampling data mixtures, retraining proxy models, and then re-performing the fitting process.
% When the components of the training dataset change (i.e., add some new datasets), previous methods like ~\citet{ye2024datamixinglawsoptimizing} and ~\citet{liu2024regmixdatamixtureregression} cannot be directly applied due to the fixed dimension of the independent variable (i.e., the number of datasets). 
% This requires resampling data mixtures, retraining proxy models, and re-performing the fitting process.
In this paper, we introduce \textsc{Domain2Vec} to decompose any dataset into a linear combination of meta-domains.
\textsc{Domain2Vec} shares some concepts with prior meta-learning works, such as~\citet{jomaa2021dataset2vec} and ~\citet{chen2024cross}, which explore dataset representation in latent spaces. 
While sharing this concept, \textsc{Domain2Vec} differs in both purpose and implementation, and we focus on the data mixture in LM pretraining.
% We also propose Distribution Alignment Assumption, a training-free method to identify the optimal data mixture for LM pretraining. 
% Notably, all fitting experiments are conducted in the domain vector space. When training datasets change, these datasets can still be mapped as linear combinations of meta-domains, significantly enhancing the scalability of prior methods~\cite{ye2024datamixinglawsoptimizing, liu2024regmixdatamixtureregression}.

% \vspace{-3pt}
\section{Conclusion}
In this work, we introduce~\textsc{Domain2Vec}, a novel method to capture the underlying features of datasets by decomposing datasets into a linear combination of several meta-domains.
It enables us to acquire vectorized representation for arbitrary datasets.
Building on these domain vectors, we introduce a training-free approach by Distribution Alignment Assumption (DA$^{2}$) to identify optimal data mixtures for language model pretraining
Furthermore, \textsc{Domain2Vec} seamlessly integrates with existing methods, greatly improving their efficiency and scalability by establishing a direct relationship between model performance and domain vectors, without requiring retraining when training datasets change.
Our experimental results demonstrate that both \textsc{Domain2Vec}+DA$^{2}$ and \textsc{Domain2Vec}+RegMix achieve comparable text generation and downstream task performance with reduced computational overhead compared to existing approaches.
We believe this work offers valuable insights into optimizing data mixtures for language model pretraining and paves the way for more efficient training strategies.

\section*{Impact Statement}
This paper presents work whose goal is to advance the field of Machine Learning. There are many potential societal consequences of our work, none which we feel must be specifically highlighted here.

\section*{Acknowledgements}
This work was supported by the National Natural Science Foundation of China (No. U24B20181) and Fujian Provincial Natural Science Foundation of China (No. 2024J08371).


\bibliography{example_paper}
\bibliographystyle{icml2025}

%%%%%%%%%%%%%%%%%%%%%%%%%%%%%%%%%%%%%%%%%%%%%%%%%%%%%%%%%%%%%%%%%%%%%%%%%%%%%%%
%%%%%%%%%%%%%%%%%%%%%%%%%%%%%%%%%%%%%%%%%%%%%%%%%%%%%%%%%%%%%%%%%%%%%%%%%%%%%%%
% APPENDIX
%%%%%%%%%%%%%%%%%%%%%%%%%%%%%%%%%%%%%%%%%%%%%%%%%%%%%%%%%%%%%%%%%%%%%%%%%%%%%%%
%%%%%%%%%%%%%%%%%%%%%%%%%%%%%%%%%%%%%%%%%%%%%%%%%%%%%%%%%%%%%%%%%%%%%%%%%%%%%%%
\newpage
\appendix
\onecolumn

% You can have as much text here as you want. The main body must be at most $8$ pages long.
% For the final version, one more page can be added.
% If you want, you can use an appendix like this one.  

% The $\mathtt{\backslash onecolumn}$ command above can be kept in place if you prefer a one-column appendix, or can be removed if you prefer a two-column appendix.  Apart from this possible change, the style (font size, spacing, margins, page numbering, etc.) should be kept the same as the main body.
%%%%%%%%%%%%%%%%%%%%%%%%%%%%%%%%%%%%%%%%%%%%%%%%%%%%%%%%%%%%%%%%%%%%%%%%%%%%%%%
%%%%%%%%%%%%%%%%%%%%%%%%%%%%%%%%%%%%%%%%%%%%%%%%%%%%%%%%%%%%%%%%%%%%%%%%%%%%%%%
\section{Detailed Description of the Distribution Alignment Assumption}
\label{appendix: Description}
In this section, we provide a detailed description of the Distribution Alignment Assumption for language model pretraining.

In the scenario of finding the optimal data mixture for language model pretraining, the validation set $\mathcal{D}_{valid}$ is fixed, and we aim to adjust the data mixture to construct the training set $\mathcal{D}_{train}$ to achieve lower validation loss calculated by Equation~\ref{eq:validation loss}, where $\hat \theta$ is parameters of a pretrained language model.
\begin{equation}
\label{eq:validation loss}
\begin{aligned}
\mathbb{E}_{X \sim \mathcal{D}_{valid}} -\log P(X | \hat \theta) =  \mathbb{E}_{X \sim \mathcal{D}_{valid}}  \sum_{i=1}^{|X|} -\log({P}(x_i| x_{<i}, \hat \theta))
\end{aligned}
\end{equation}
Typically, we pretrain language models via next token prediction~\cite{radford2018improving} like Equation~\ref{eq:next token prediction}.
\begin{align}
\label{eq:next token prediction}
\begin{aligned}
\hat \theta &= \arg\max_{\theta} \mathbb{E}_{X \sim \mathcal{D}_{train}} \log P(X|\theta) \\ 
&= \arg\max_{\theta} \mathbb{E}_{X \sim \mathcal{D}_{train}}  \sum_{i=1}^{|X|} \log({P}(x_i| x_{<i}, \theta))
\end{aligned}
\end{align}
That is, we need to find a $\hat \theta$ that maximizes the expected probability of $X \sim \mathcal{D}_{train}$, which is also known as Maximum Likelihood Estimation (MLE).
When the data distributions of $\mathcal{D}_{\text{train}}$ and $\mathcal{D}_{\text{valid}}$ are aligned, the optimization target of language models pretraining process equals find a $\hat \theta$ that maximizes the expected probability of $X \sim \mathcal{D}_{valid}$.
Therefore, we introduce the Distribution Alignment Assumption for language model pretraining, a novel method to find the optimal data mixture without training.

\section{Algorithm}
\label{appendix: Algorithm}
In Algorithm~\ref{alg: domain2vec}, we present pseudo code for acquiring the domain vectors of training and validation datasets.

In Algorithm~\ref{alg: predict_the_data_mixture_daa} and ~\ref{alg: predict_the_data_mixture_regmix}, we present pseudo code for how to use \textsc{Domain2Vec} to find the optimal data mixture, including Distribution Alignment Assumption, and applying \textsc{Domain2Vec} to RegMix.

Note that when applying \textsc{Domain2Vec}+DA$^2$ or \textsc{Domain2Vec}+RegMix , for getting more stable and accurate results, one could also average the $k$-best ratios in the K sampled candidates data mixture. We present top-$1$ as one example in the pseudo codes. We adopt top-$1$ for direct comparison in Section~\ref{sec:task1}, while we adopt top-$100$ in Section~\ref{sec:task2}, which is aligned with RegMix~\cite{liu2024regmixdatamixtureregression}.

\begin{algorithm}[!ht]
\caption{\textsc{Domain2Vec}}
\label{alg: domain2vec}
\begin{algorithmic}[1]
\REQUIRE Training datasets $\mathcal D_{train} = \{ \mathcal D_1, \mathcal D_2,...,\mathcal D_m \}$ , validation dataset $\mathcal{D}_{valid}$, meta-domain classifier $\mathrm{Classifier}$
\\
\STATE Domain vectors $V_{train} = []$
\FOR{$i = 1$ \textbf{to} $m$}
    \STATE Sample $N$ documents from $\mathcal{D}_i$
    \STATE $\boldsymbol v_i = \frac{1}{N}\sum_{j=1}^{N} \mathrm{Classifier}(doc_j)$, where $doc_j \in \mathcal{D}_i $ 
\ENDFOR
\\
\STATE Sample $N$ documents from $\mathcal{D}_{valid}$
\STATE $\boldsymbol v_{valid} = \frac{1}{N}\sum_{j=1}^{N} \mathrm{Classifier}(doc_j)$, where $doc_j \in \mathcal{D}_{valid}$
\\
\STATE {\bfseries Return:} $\boldsymbol V_{train} = [\bm{v}_1, \bm{v}_2,...,\bm{v}_m ], \bm{v}_{valid}$
\end{algorithmic}
\end{algorithm}

\begin{algorithm}[!t]
\caption{\textsc{Domain2Vec}+DA$^2$}
\label{alg: predict_the_data_mixture_daa}
\begin{algorithmic}[1]
\REQUIRE Domain vectors of training datasets $\boldsymbol V_{train} = [\boldsymbol v_1, \boldsymbol v_2,...,\boldsymbol v_m ]$, domain vectors of validation dataset  $\boldsymbol v_{valid}$, token distribution of training datasets $\boldsymbol a_{train}$.
\\
\STATE Sample $K$ candidates data mixture $\boldsymbol r_i$ from $\mathrm{Dirichlet}(\boldsymbol a_{train})$ 
\\
\STATE The optimal data mixture $\boldsymbol {r}^* = \boldsymbol r_1$ 
\\
\FOR{$i = 2$ {\bfseries to} $K$}  
    \IF{$\mathrm{Dist}(\boldsymbol V_{train}\cdot \boldsymbol r, \boldsymbol v_{valid}) <  \mathrm{Dist}(\boldsymbol V_{train}\cdot \boldsymbol r^*, \boldsymbol v_{valid})$} 
    % \COMMENT{Updata the optimal data mixture} 
    \STATE $\boldsymbol r^* = \boldsymbol r_i$
    \ENDIF
\ENDFOR
\\
\STATE {\bfseries Return:} the optimal data mixture $\boldsymbol r^* $
\end{algorithmic}
\end{algorithm}

\begin{algorithm}[!t]
\caption{\textsc{Domain2Vec}+RegMix}
\label{alg: predict_the_data_mixture_regmix}
\begin{algorithmic}[1]
\REQUIRE Domain vectors of training datasets $\boldsymbol V_{train} = [\boldsymbol v_1, \boldsymbol v_2,\cdots, \boldsymbol v_m ]$, domain vectors of validation dataset $\boldsymbol v_{valid} = [q_1, q_2, \cdots, q_n]^\top$, token distribution of training datasets $\boldsymbol a_{train}$, fitted model for each meta-domain $f_i(\cdot)$.
\\
\STATE Sample $K$ candidates data mixture $\boldsymbol r_i$ from $\mathrm{Dirichlet}(\boldsymbol a_{train})$ 
\STATE The optimal data mixture $r^* = r_1$ 
% \STATE $\mathcal{L}(\boldsymbol r^*) = \sum_{i=1}^{n} q_i \cdot \mathcal L^{\mathcal D_{i}^*}(\boldsymbol V_{train}\cdot \boldsymbol r_1)$
\STATE Def $\mathcal L(\bm{r}) =  \sum \limits_{i=1}^{n} q_i \cdot f_i(\bm V_{train}\cdot\bm{r})$
\FOR{$i = 2$ {\bfseries to} $K$}    
    \IF{$\mathcal{L}(\boldsymbol r_i) < \mathcal{L}(\boldsymbol r^*)$} 
    \STATE $\boldsymbol r^* = \boldsymbol r_i$
    \STATE $\mathcal{L}(\boldsymbol r^*) = \mathcal{L}(\boldsymbol r_i)$
    \ENDIF
\ENDFOR
\\
\STATE {\bfseries Return:} the optimal data mixture $\boldsymbol r^* $
\end{algorithmic}
\end{algorithm}

\section{Data Mixture of Different Methods}
In this section, we will show the data mixture on The Pile~\citep{DBLP:journals/corr/abs-2101-00027} of different methods we used in this paper for reproduction.
In Table~\ref{table: data mixture of pile}, we show the optimal data mixture predicted by \textsc{Domain2Vec} + DA$^2$ and  \textsc{Domain2Vec} + RegMix. 
It should be noted that, to avoid the over-fitting problem, any subset of The Pile~\citep{DBLP:journals/corr/abs-2101-00027} will be only trained at most one epoch. 
Because we adopt rejection sampling to filter out certain unreasonable data mixtures. 
The data mixture predicted may change as model sizes change.

\section{Experimental Results of Pilot Study}
\label{appendix: additional results}
\begin{figure}[!t]
%\vspace{-1em}
    \centering
    \includegraphics[width=0.9\textwidth]{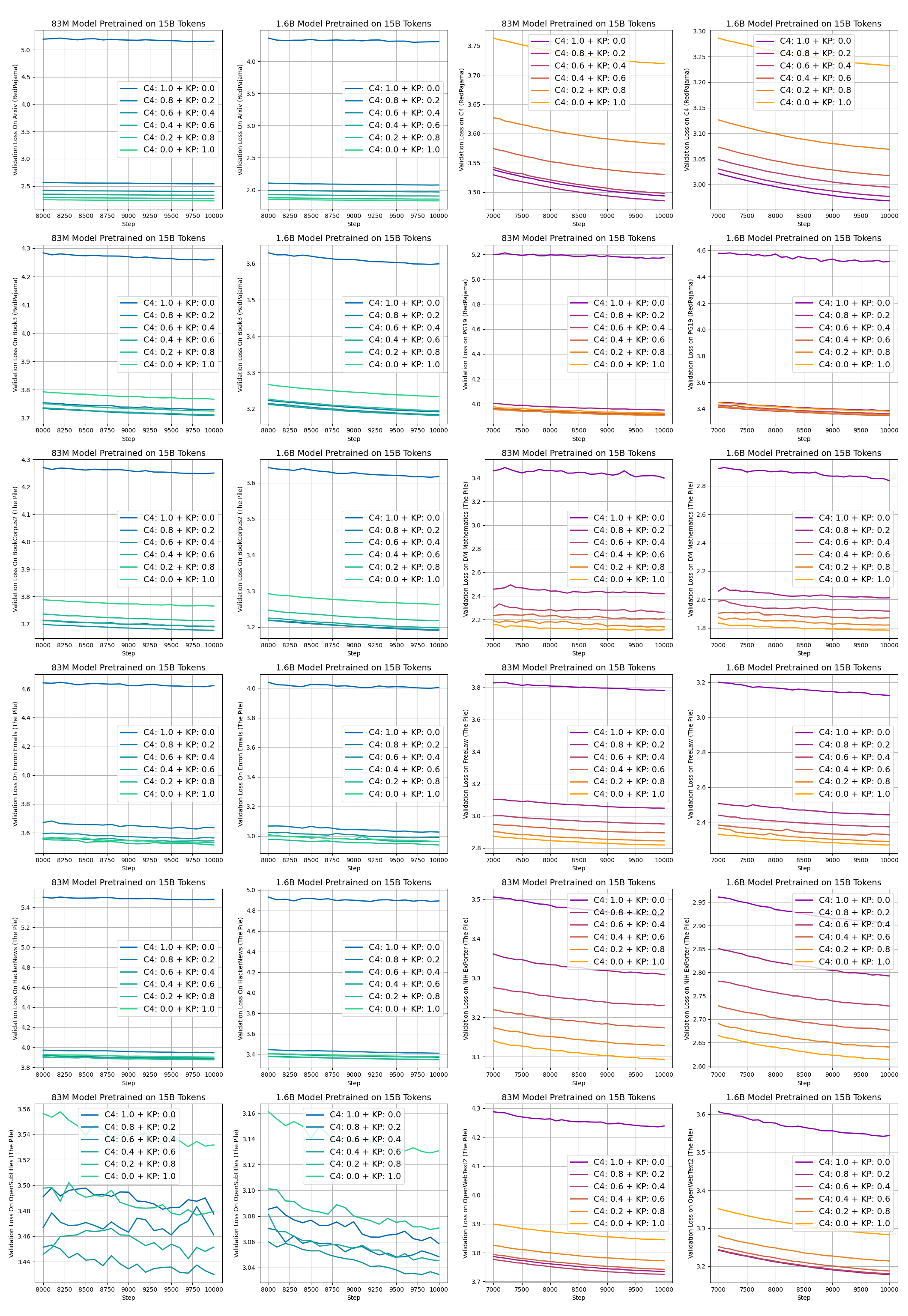}
    %\vspace{-1.5em}
    \caption{The validation loss on different dataset of models trained using data mixture in Table~\ref{table: ood data mixture}.}
    \label{fig:pilot study 1}
    %\vspace{-1.0em}
\end{figure}

\begin{figure}[!t]
%\vspace{-1em}
    \centering
    \includegraphics[width=0.9\textwidth]{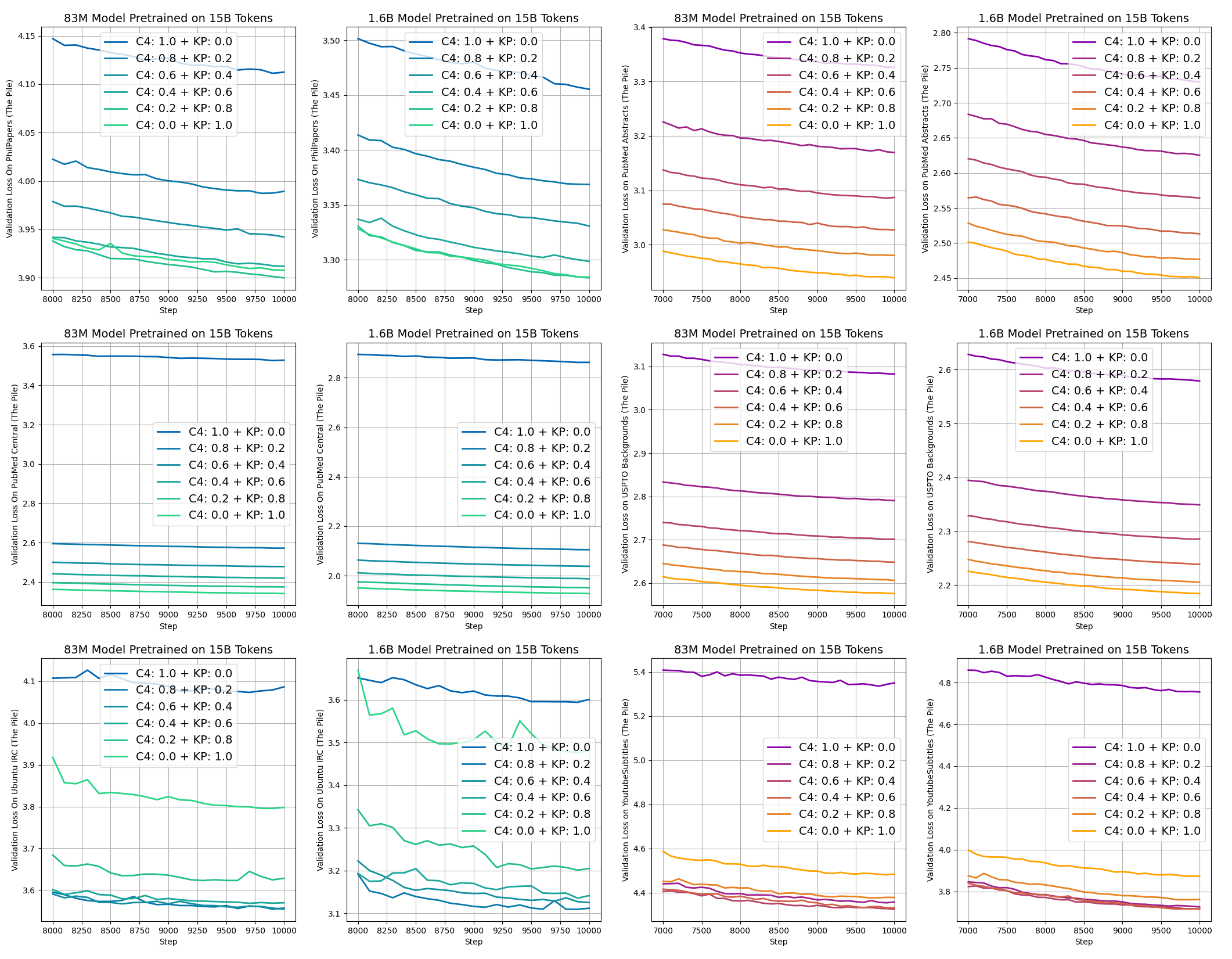}
    %\vspace{-1.5em}
    \caption{The validation loss on different dataset of models trained using data mixture in Table~\ref{table: ood data mixture}. }
    \label{fig:pilot study 2}
    %\vspace{-1.0em}
\end{figure}

In this section, we report the validation loss on various datasets arXiv, C4, Book3, PG19 from RedPajama~\citep{together2023redpajama}, and BookCorpus2, DM Mathematics, Enron Emails, FreeLaw, HackerNews, NIH ExPorter, OpenSubtitles, OpenWebText2, PhilPapers, PubMed Abstracts, PubMed Central, USPTO Backgrounds, Ubuntu IRC, Youtube Subtitles from The Pile~\citep{DBLP:journals/corr/abs-2101-00027} in Figure~\ref{fig:pilot study}, Figure~\ref{fig:pilot study 1} and Figure~\ref{fig:pilot study 2}. 
% According to the experimental results, we find that 1) \emph{\textbf{for different validation sets, the ranking of mixture ratios varies significantly}}. 2) \emph{\textbf{for the same validation set, the data mixture ranking of validation loss on identical validation dataset does not change with the variation in model parameters}}. 
% We hope our experimental results and findings could provide some insights to the community about efficiently finding the optimal data mixture.

\section{Comparative Study on Different Distributional Measures of DA$^2$}
\label{appdenix: Different Distributional Measures}
In Section~\ref{sec: da2}, we use Huber Loss to measure the similarity of domain vectors.
Technically, Huber loss combines the advantages of L1 and L2 distance. 
In Table~\ref{table: Different Distributional Measures}, we add the results of different distributional measures. 
As shown in the Table~\ref{table: Different Distributional Measures}, Huber Loss shows better performance than L1/L2/JS Distance.
Additionally, Wasserstein distance is a very great option.
However, it would require an extra metric space matrix, $\boldsymbol{M}$, to measure the distance between two domain vectors. 
In this work, the metric space, $\boldsymbol{M} \in \mathbb R^{260 \times 260}$, is actually the ``dataset transition cost'' between each two meta-domains, and is non-trivial. 
Each element in $\boldsymbol{M}$. $c_{i,j}$ could be estimated via $\mathcal{L}_{i,j}$ , the loss at meta-domain $j$ after training on meta-domain $i$, which requires additional computational resources. 
Considering that Huber Loss already achieved very positive results, we did not conduct this experiment. 
We believe that Wasserstein distance can also present a positive result (even better) if the metric space is well estimated, and we leave this for future work.

\newpage

\begin{table*}[!t]
\caption{Huber Loss shows better performance than L1/L2/JS Distance.}
\label{table: Different Distributional Measures}
\vskip 0.15in
\begin{center}
\footnotesize
\resizebox{0.45\textwidth}{!}{
\begin{tabular}{l|cc}
\toprule
\multirow{1}{*}{\textbf{Distributional Measure}} &
\multicolumn{1}{c}{\textbf{Pearson} }  &
\multicolumn{1}{c}{\textbf{Spearman}} \\                          
\midrule
Huber Loss  & 0.5833 & 0.6657 \\
JS Distance & 0.4527 & 0.5000 \\
L1 Distance & 0.4830 & 0.5400 \\
L2 Distance & 0.5720 & 0.6429 \\
\bottomrule
\end{tabular}
}
\end{center}
\end{table*}

\begin{table*}[!t]
\caption{The data mixture of The Pile~\citep{DBLP:journals/corr/abs-2101-00027} from different baselines, which aligns with the data mixture used in~\citet{liu2024regmixdatamixtureregression}.}
\label{table: data mixture of pile}
\vskip 0.15in
\begin{center}
\footnotesize
\resizebox{0.65\textwidth}{!}{
\begin{tabular}{l|cccc}
\toprule
\multirow{1}{*}{\textbf{Data Mixture}} &
\multicolumn{1}{c}{\textbf{Human} }  &
\multicolumn{1}{c}{\textbf{DoReMi} }  &
\multicolumn{1}{c}{\textbf{Pile-CC Only} }  &
\multicolumn{1}{c}{\textbf{RegMix} }  
\\                          
\midrule
ArXiv & 0.134 & 0.004 & 0.0 & 0.001  \\
FreeLaw & 0.049 & 0.005 & 0.0 & 0.001 \\
NIH ExPorter & 0.007 & 0.008  &  0.0  & 0.001 \\
PubMed Central & 0.136 & 0.006 & 0.0 & 0.003  \\
Wikipedia (en) & 0.117 & 0.086 & 0.0 & 0.016 \\
DM Mathematics & 0.025 & 0.002 & 0.0 & 0.0 \\
Github  & 0.054 & 0.022 & 0.0 & 0.0 \\
PhilPapers & 0.003 & 0.034 & 0.0 & 0.0 \\
Stack Exchange & 0.118 & 0.019 & 0.0 & 0.0 \\
Enron Emails  & 0.004 & 0.009 & 0.0 & 0.002 \\
Gutenberg (PG-19) & 0.025 & 0.009 & 0.0 & 0.002 \\
Pile-CC & 0.142 & 0.743 & 1.0 & 0.87  \\
Ubuntu IRC & 0.009 & 0.011 & 0.0 & 0.064 \\
EuroParl & 0.005 & 0.008 & 0.0 & 0.0 \\
HackerNews & 0.01 & 0.016 & 0.0 & 0.012  \\
PubMed Abstracts & 0.107 & 0.014 & 0.0 & 0.024 \\
USPTO Backgrounds & 0.053 & 0.004 & 0.0 & 0.002 \\
\bottomrule
\end{tabular}
}
\end{center}
\end{table*}

\begin{table*}[!t]
\caption{The optimal data mixture predicted by \textsc{Domain2Vec} + DA$^2$ and  \textsc{Domain2Vec} + RegMix. To avoid the over-fitting problem, any subset of The Pile~\citep{DBLP:journals/corr/abs-2101-00027} will be trained at most one epoch. And we adopt rejection sampling to filter out certain unreasonable data mixtures. Thus, the data mixture predicted may change as model sizes change.}
\label{table: our data mixture of pile}
\vskip 0.15in
\begin{center}
\footnotesize
\resizebox{0.75\textwidth}{!}{
\begin{tabular}{l|cccc|cccc}
\toprule
\multicolumn{1}{l|}{\textbf{Data Mixture} } & \multicolumn{4}{c|}{\textbf{\textsc{Domain2Vec}+DA$^2$}} & \multicolumn{4}{c}{\textbf{\textsc{Domain2Vec}+RegMix}}
\\
\cmidrule(lr){2-5}  \cmidrule(lr){6-9}  
&
\multicolumn{1}{c}{\textbf{106M} }  &
\multicolumn{1}{c}{\textbf{290M} }  &
\multicolumn{1}{c}{\textbf{595M} }  &
\multicolumn{1}{c|}{\textbf{1B}  }  &
\multicolumn{1}{c}{\textbf{106M} }  &
\multicolumn{1}{c}{\textbf{290M} }  &
\multicolumn{1}{c}{\textbf{595M} }  &
\multicolumn{1}{c}{\textbf{1B}   } 
\\                          
\midrule
ArXiv              & 0.0131 & 0.0131  & 0.0389 & 0.0431 & 0.0152 & 0.0070 & 0.0114 & 0.0103\\
FreeLaw            & 0.0076 & 0.0076  & 0.0316 & 0.0305 & 0.0395 & 0.0267 & 0.0339 & 0.0268\\
NIH ExPorter       & 0.0008 & 0.0008  & 0.0028 & 0.0023 & 0.0000 & 0.0199 & 0.0000 & 0.0000\\
PubMed Central     & 0.0773 & 0.0773  & 0.0519 & 0.0704 & 0.0343 & 0.0576 & 0.0099 & 0.0518\\
Wikipedia (en)     & 0.2970 & 0.2970  & 0.2049 & 0.2126 & 0.0847 & 0.0101 & 0.1014 & 0.2577\\ 
DM Mathematics     & 0.0003 & 0.0003  & 0.0056 & 0.0026 & 0.0177 & 0.0018 & 0.0011 & 0.0008\\
Github             & 0.0096 & 0.0096  & 0.0290 & 0.0298 & 0.0034 & 0.0538 & 0.0500 & 0.0138\\
PhilPapers         & 0.0018  & 0.0018  & 0.0093 & 0.0025 & 0.0118 & 0.0005 & 0.0333 & 0.0401\\
Stack Exchange     & 0.0464 & 0.0464  & 0.0661 & 0.0585 & 0.0698 & 0.0430 & 0.1199 & 0.0262\\
Enron Emails       & 0.0000 & 0.0000  & 0.0009 & 0.0000 & 0.0018 & 0.0000 & 0.0000 & 0.0000\\
Gutenberg (PG-19)  & 0.0217 & 0.0217  & 0.0484 & 0.0370 & 0.0467 & 0.0223 & 0.0007 & 0.0252\\
Pile-CC            & 0.4338 & 0.4338  & 0.3191 & 0.3814 & 0.5370 & 0.6323 & 0.5546 & 0.4704\\
Ubuntu IRC         & 0.0022 & 0.0022  & 0.0063 & 0.0072 & 0.1019 & 0.0123 & 0.0161 & 0.0069\\
EuroParl           & 0.0003 & 0.0003  & 0.0042 & 0.0040 & 0.0070 & 0.0037 & 0.0116 & 0.0000\\
HackerNews         & 0.0154 & 0.0154  & 0.0521 & 0.0199 & 0.0028 & 0.0551 & 0.0170 & 0.0673\\
PubMed Abstracts   & 0.0596 & 0.0596  & 0.0739 & 0.0532 & 0.0259 & 0.0102 & 0.0190 & 0.0017\\
USPTO Backgrounds  & 0.0130 & 0.0130  & 0.0549 & 0.0449 & 0.0004 & 0.0438 & 0.0201 & 0.0010\\
\bottomrule
\end{tabular}
}
\end{center}
\end{table*}

\begin{table*}[!t]
\caption{The parameters of different models we used in Section~\ref{sec:task1} and Section~\ref{sec:task2}. When calculating the model parameters, we do not take into account the embedding layer and the language model head layer.}
\label{table: model parameters}
\vskip 0.15in
\begin{center}
\footnotesize
\resizebox{0.7\textwidth}{!}{
\begin{tabular}{l|cc|cccc}
\toprule
\multicolumn{1}{l|}{\textbf{Parameter} } & \multicolumn{2}{c|}{\textbf{Text Generation}} & \multicolumn{4}{c}{\textbf{Downstream Task}}
\\
\cmidrule(lr){2-3}  \cmidrule(lr){4-7}  
&
\multicolumn{1}{c}{\textbf{83M} }  &
\multicolumn{1}{c}{\textbf{1.6B} }  &
\multicolumn{1}{c}{\textbf{106M} }  &
\multicolumn{1}{c}{\textbf{290M} }  &
\multicolumn{1}{c}{\textbf{595M} }  &
\multicolumn{1}{c}{\textbf{1B}   } 
\\                          
\midrule
Hidden Size        & 768    & 2,048    & 768    & 1,280   & 1,536   & 2,048  \\
FFN Hidden Size    & 2,048   & 5,504    & 2,048   & 3,392   & 4,096   & 5,440  \\
Num of Layers      & 12     & 24      & 15     & 15     & 21     & 21    \\
Num of Heads       & 12     & 16      & 12     & 10     & 12     & 32    \\
Max Seq Length     & 4,096   & 4,096    & 4,096   & 4,096   & 4,096   & 4,096  \\ 
Vocab Size         & 128,256 & 128,256  & 151,936 & 151,936 & 151,936 & 151,936\\
RoPE Base          & 10,000  & 10,000   & 10,000  & 10,000  & 10,000  & 10,000 \\
\bottomrule
\end{tabular}
}
\end{center}
\end{table*}

\begin{table*}[!t]
%\vspace{-4.5mm}
\caption{Downstream Task Performance of different data mixture on 106M Model. Similar to~\citet{liu2024regmixdatamixtureregression}, Human refers the original data mixture from The Pile. Pile-CC Only refers only training on the Pile-CC subset. The data mixture and estimated flops of DoReMi and RegMix are from~\citet{liu2024regmixdatamixtureregression}.}
\label{table: 106M models' main results}
\begin{center}
\footnotesize
\resizebox{0.98\textwidth}{!}{
\begin{tabular}{l|cccccc}
\toprule
\multicolumn{1}{l}{\textbf{Benchmark} }  &
\multicolumn{1}{c}{\textbf{Human} }  &
\multicolumn{1}{c}{\textbf{DoReMi} }  &
\multicolumn{1}{c}{\textbf{Pile-CC Only} }  &
\multicolumn{1}{c}{\textbf{RegMix} }  &
\multicolumn{1}{c}{\textbf{\textsc{Domain2Vec} $+$ DA$^{2}$} }  &
\multicolumn{1}{c}{\textbf{\textsc{Domain2Vec} $+$ RegMix} } 
\\                          
\midrule
\multicolumn{7}{c}{\emph{106M Model Pretrained on 2B Tokens}} \\
\midrule 	 	 	 	 	
Social IQA     & 0.340 & 0.349 & 0.353 & 0.356 & 0.339 & 0.342 \\
HellaSwag      & 0.268 & 0.268 & 0.269 & 0.269 & 0.267 & 0.264 \\
PiQA           & 0.573 & 0.584 & 0.580 & 0.586 & 0.579 & 0.583 \\
OpenBookQA     & 0.245 & 0.251 & 0.249 & 0.242 & 0.245 & 0.249 \\
Lambada        & 0.065 & 0.099 & 0.102 & 0.091 & 0.091 & 0.090 \\
SciQ           & 0.550 & 0.520 & 0.509 & 0.537 & 0.549 & 0.518 \\
ARC Easy       & 0.329 & 0.339 & 0.335 & 0.337 & 0.334 & 0.331 \\
COPA           & 0.525 & 0.570 & 0.572 & 0.585 & 0.578 & 0.557 \\
RACE           & 0.236 & 0.254 & 0.246 & 0.251 & 0.240 & 0.244 \\
LogiQA         & 0.282 & 0.280 & 0.271 & 0.274 & 0.268 & 0.286 \\
WinoGrande     & 0.516 & 0.516 & 0.502 & 0.508 & 0.506 & 0.499 \\
MultiRC        & 0.539 & 0.520 & 0.515 & 0.533 & 0.541 & 0.544 \\
\rowcolor{gray!20}
Average Performance & 0.372 & 0.379 & 0.375 & 0.381 & 0.378 & 0.376 \\
\midrule
 \multicolumn{7}{c}{\emph{290M Model Pretrained on 6B Tokens}} \\
\midrule 	 	 	 	 	
Social IQA  & 0.364 & 0.373 & 0.374 & 0.371 & 0.371 & 0.368  \\
HellaSwag & 0.295 & 0.312 & 0.317 & 0.315 & 0.307 & 0.312 \\
PiQA  & 0.605 & 0.631 & 0.639 & 0.642 & 0.624 & 0.633 \\
OpenBookQA & 0.261 & 0.271 & 0.271 & 0.262 & 0.268 & 0.266 \\
Lambada & 0.175 & 0.208 & 0.206 & 0.210 & 0.182 & 0.208 \\
SciQ  & 0.711 & 0.682 & 0.663 & 0.674 & 0.670 & 0.697 \\
ARC Easy & 0.395 & 0.410 & 0.419 & 0.417 & 0.420 & 0.412 \\
COPA & 0.632 & 0.660 & 0.682 & 0.657 & 0.627 & 0.642 \\
RACE & 0.265 & 0.280 & 0.280 & 0.276 & 0.283 & 0.281\\
LogiQA &0.283 & 0.293 & 0.296 & 0.276 & 0.277 & 0.292 \\
WinoGrande & 0.511 & 0.506 & 0.509 & 0.524 & 0.498 & 0.504 \\
MultiRC & 0.507 & 0.555 & 0.513 & 0.545 & 0.521 & 0.517 \\
\rowcolor{gray!20}
Average Performance & 0.417 & 0.432 & 0.431 & 0.431 & 0.421 & 0.428 \\
\midrule
\multicolumn{7}{c}{\emph{595M Model Pretrained on 12B Tokens}} \\
\midrule
Social IQA  & 0.378 & 0.387 & 0.390 & 0.394 & 0.383 & 0.388 \\
HellaSwag & 0.338 & 0.377 & 0.386 & 0.385 & 0.355 & 0.366 \\
PiQA  & 0.624 & 0.656 & 0.663 & 0.667 & 0.651 & 0.659 \\
OpenBookQA & 0.273 & 0.279 & 0.283 & 0.294 & 0.288 & 0.271 \\
Lambada & 0.255 & 0.294 & 0.332 & 0.310 & 0.269 & 0.292 \\
SciQ  & 0.777 & 0.757 & 0.770 & 0.791 & 0.763 & 0.769\\
ARC Easy & 0.439 & 0.453 & 0.478 & 0.481 & 0.453 & 0.460 \\
COPA & 0.642 & 0.680 & 0.672 & 0.663 & 0.668 & 0.667 \\
RACE & 0.289 & 0.309 & 0.311 & 0.311 & 0.288 & 0.303\\
LogiQA & 0.263 & 0.268 & 0.252 & 0.267 & 0.263 & 0.267\\
WinoGrande & 0.509 & 0.515 & 0.506 & 0.509 & 0.512 & 0.503\\
MultiRC & 0.516 & 0.533 & 0.522 & 0.507 & 0.506 & 0.527\\
\rowcolor{gray!20}
Average Performance & 0.442 & 0.459 & 0.464 & 0.465 & 0.450 & 0.456 \\
\midrule
\multicolumn{7}{c}{\emph{1B Model Pretrained on 20B Tokens}} \\
\midrule
Social IQA  & 0.387 & 0.411 & 0.406 & 0.406 & 0.394 & 0.401 \\
HellaSwag & 0.375 & 0.427 & 0.431 & 0.436 & 0.410 & 0.410\\
PiQA  & 0.658 & 0.684 & 0.693 & 0.691 & 0.684 & 0.680 \\
OpenBookQA & 0.278 & 0.298 & 0.300 & 0.304 & 0.299 & 0.302\\
Lambada & 0.301 & 0.359 & 0.348 & 0.353 & 0.334 & 0.339 \\
SciQ  & 0.802 & 0.822 & 0.809 & 0.828 & 0.821 & 0.818 \\
ARC Easy & 0.482 & 0.508 & 0.512 & 0.518 & 0.500 & 0.499 \\
COPA & 0.683 & 0.692 & 0.713 & 0.708 & 0.678 & 0.698 \\
RACE & 0.306 & 0.319 & 0.313 & 0.314 & 0.305 & 0.300\\
LogiQA & 0.259 & 0.258 & 0.269 & 0.272 & 0.268 & 0.267 \\
WinoGrande & 0.513 & 0.527 & 0.541 & 0.512 & 0.535 & 0.533 \\
MultiRC & 0.523 & 0.504 & 0.510 & 0.530 & 0.529 & 0.548 \\
\rowcolor{gray!20}
Average Performance & 0.464 & 0.484 & 0.487 & 0.489 & 0.480 & 0.483\\
 \midrule
\multirow{2}{*}{Estimated FLOPs}
 & \multirow{2}{*}{$0$}
 & $3.7 \times 10^{19} $ & \multirow{2}{*}{$0$} & $3.5 \times 10^{18}$ & $9.66 \times 10^{16}$  & $9.66 \times 10^{16}$ \\
 &  & $(100\%)$ &  &  $(9.46\%)$ &  $(0.26\%)$ &  $(0.26\%)$ \\ 

\bottomrule
\end{tabular}
}
%\vspace{-1em}
\end{center}
\end{table*}

\newpage

\begin{figure}[!ht]
%\vspace{-1em}
    \centering
    \includegraphics[width=1.0\textwidth]{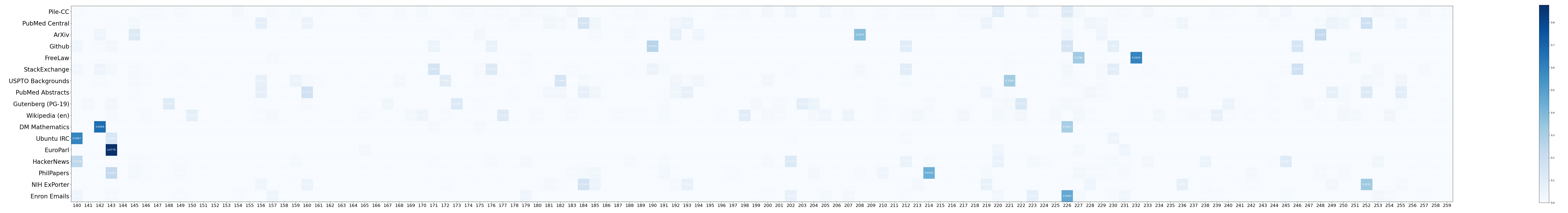}
    %\vspace{-7.0mm}
    \caption{The Domain Vector of each sub-dataset of The Pile~\citep{DBLP:journals/corr/abs-2101-00027}, where each row corresponds to a sub-dataset and each column corresponds to a meta-domain. The higher the proportion of data belonging to a particular meta-domain, the closer the color of the corresponding cell is to \textcolor[RGB]{0,0,255}{blue}). Additionally, since The Pile primarily consists of English texts, we only display the distribution on English meta-domains for clarity.}
    \label{fig:pile_en_full}
\end{figure}
%\vspace{-1em}

\newpage

\end{document}